%% file: sigplan.tex
\documentclass[sigplan,10pt]{acmart}

\usepackage[english]{babel}
\usepackage{blindtext}
\usepackage{pifont}  
\usepackage{bbding} 
\usepackage{xcolor}[table]
\usepackage{colortbl}
\usepackage{subfigure}
\usepackage{multirow}
\usepackage{booktabs}
\usepackage{graphicx}
\AtBeginDocument{%
  }

\copyrightyear{2026}\acmYear{2026}
\setcopyright{cc}
\setcctype{by-nc-nd}
\acmConference[SIGOPS ATC 2026]{The 2026 ACM SIGOPS Annual Technical Conference}{November 15-18, 2026}{Hong Kong, China}
\acmBooktitle{The 2026 ACM SIGOPS Annual Technical Conference (SIGOPS ATC 2026), November 15-18, 2026, Hong Kong, China}
\acmDOI{10.1145/3828523.3857118}
\acmISBN{979-8-4007-2818-1/2026/11}

\begin{document}

\pagestyle{empty}
\title{FlexMARL: Graph-Driven Rollout-Training Co-Design for LLM-Based Multi-Agent Reinforcement Learning}

\settopmatter{authorsperrow=4}
\author{Zhida Jiang}
\affiliation{%
  \institution{JD.com}
  \city{Beijing}
  \country{China}
}
\email{zdjiang98@gmail.com}

\author{Zhaolong Xing}
\affiliation{%
  \institution{JD.com}
  \city{Beijing}
  \country{China}
}
\email{nhzlx.dragon@gmail.com}

\author{Jiawei Lu}
\affiliation{%
  \institution{JD.com}
  \city{Beijing}
  \country{China}
}
\email{lujiawei30@jd.com}

\author{Yipei Niu}
\affiliation{%
  \institution{Huawei}
  \city{Shanghai}
  \country{China}
}
\email{newypei@gmail.com}

\author{Qingyuan Sang}
\affiliation{%
  \institution{JD.com}
  \city{Beijing}
  \country{China}
}
\email{sangqingyuan1@jd.com}

\author{Liangxu Zhang}
\affiliation{%
  \institution{JD.com}
  \city{Beijing}
  \country{China}
}
\email{zhangliangxu3@jd.com}

\author{Wenquan Dai}
\affiliation{%
  \institution{Huawei}
  \city{Shanghai}
  \country{China}
}
\email{daiwenquan1@huawei.com}

\author{Junhua Shu}
\affiliation{%
  \institution{JD.com}
  \city{Beijing}
  \country{China}
}
\email{shujunhua1@jd.com}

\author{Jiaxing Wang}
\affiliation{%
  \institution{JD.com}
  \city{Beijing}
  \country{China}
}
\email{wjiaxing94@gmail.com}

\author{Qiangyu Pei}
\affiliation{%
  \institution{Huawei}
  \city{Shanghai}
  \country{China}
}
\email{peiqiangyu@huawei.com}

\author{Qiong Chen}
\affiliation{%
  \institution{Huawei}
  \city{Hangzhou}
  \country{China}
}
\email{chenqiong13@huawei.com}

\author{Xinyu Liu}
\affiliation{%
  \institution{Huawei}
  \city{Shanghai}
  \country{China}
}
\email{liuxinyu52@huawei.com}

\author{Fangming Liu}
\authornote{Corresponding Authors.}
\affiliation{%
  \institution{Huazhong University of Science and Technology}
  \city{Wuhan}
  \country{China}
}
\email{fangminghk@gmail.com}

\author{Ai Han}
\affiliation{%
  \institution{JD.com}
  \city{Beijing}
  \country{China}
}
\email{hanai@amss.ac.cn}

\author{Zhen Chen}
\authornotemark[1]
\affiliation{%
  \institution{JD.com}
  \city{Beijing}
  \country{China}
}
\email{chenzhen48@jd.com}

\author{Ke Zhang}
\affiliation{%
  \institution{JD.com}
  \city{Beijing}
  \country{China}
}
\email{wang\_ke@126.com}

\renewcommand{\shortauthors}{Zhida Jiang et al.}

\begin{abstract}
Despite algorithm-level innovations for multi-agent reinforcement learning (MARL), the underlying infrastructure for large-scale MARL remains underexplored. Existing training frameworks primarily optimize for single-agent scenarios and fail to capture the graph-structured nature of MARL. This fundamental mismatch leads to three MARL-specific challenges: topology-induced rollout skew, critical path synchronization barriers, and dynamic training demand.
To bridge this gap, we propose FlexMARL, a graph-driven rollout-training co-designed framework for large-scale LLM-based MARL.\enlargethispage{20pt} 
FlexMARL introduces the multi-agent execution graph (MEG) and materializes the intermediate representation as the \textit{MEG Store}, which provides decentralized multi-table storage and serves as the data hub for system components.
Built around the MEG Store, the \textit{Rollout Engine} develops dependency parallel sampling with hierarchical load balancing, which reduces inference latency and balances intra-agent/inter-agent loads.
The \textit{Joint Orchestrator} exploits an agent-level asynchronous pipeline to mitigate the critical path synchronization barriers while maintaining consistency. 
The \textit{Training Engine} achieves on-demand resource allocation by collectively managing training processes for individual agents. The training states of different agents are swapped via location-agnostic communication APIs.
Empirical results on a production cluster demonstrate that FlexMARL achieves up to 7.3$\times$ speedup and improves hardware utilization by up to 5.6$\times$ compared to existing frameworks. \enlargethispage{20pt}
\end{abstract}




\maketitle

\section{Introduction}
\input{content/intro}

\section{Background and Motivation}\label{sec:Background}
\input{content/background}

\section{System Overview}

\input{content/overview}

\section{MEG Store}\label{sec:MEG}
\input{content/graph_abstraction}

\section{Rollout Engine}\label{sec:Load-Aware Rollout Engine}

\input{content/rollout_engine}

\section{Joint Orchestrator}\label{sec:Joint Orchestrator}

\input{content/orchestrator}

\section{Training Engine}\label{sec:Resource-Efficient Training Engine}

\input{content/training_engine}

\section{Implementation and Deployment}\label{sec:implementation}
\input{content/implementation}

\section{Performance Evaluation}\label{sec:Evaluation}
\input{content/evaluation}

\section{Related Work}

\input{content/related_work}

\section{Conclusion}
In this paper, we present FlexMARL, a rollout-training co-designed framework that bridges the gap between LLM-based MARL algorithms and infrastructure. 
We identify three MARL-specific challenges that hinder large-scale training efficiency.
Building upon the MEG Store, FlexMARL designs dependency parallel sampling with load balancing for rollout, agent-level asynchronous pipeline for orchestration, and on-demand resource allocation for training.
The evaluation results on a production cluster confirm the superior scalability and efficiency of our proposed framework.




\bibliographystyle{ACM-Reference-Format}
\bibliography{reference}


\end{document}

%% file: content/intro.tex
Recent advances in large language models (LLMs) have facilitated the development of multi-agent systems that leverage specialized roles and distributed reasoning to solve complex tasks \cite{li2024survey,wu2024autogen,hong2023metagpt}. 
Concurrently, reinforcement learning (RL) has emerged as a powerful paradigm for improving the training performance of LLMs \cite{guo2025deepseek,yu2025dapo}. 
To harness RL's policy optimization capabilities for strengthening multi-agent systems, multi-agent reinforcement learning (MARL) enables a collective of agents to learn coordinated policies through shared experiences \cite{sun2024llm,yu2022surprising}.
Such a learning paradigm continuously collects experience generated by multi-agent interactions, integrates feedback from reward signals, and refines models to unleash collaborative intelligence \cite{zhao2025stronger,marti2025,han2025joyagents}.

Despite the impressive performance gains brought by MARL, most research pays more attention to algorithm-level innovations, such as communication protocol \cite{zhang2025learning,wittner2026communication}, coordination optimization \cite{qin2025strategic,liu2026learning}, and reward function refinement \cite{wei2025lero,carter2026multi}.
The underlying networked infrastructure that supports large-scale MARL training remains underexplored.  
Mainstream RL frameworks (e.g., OpenRLHF \cite{hu2024openrlhf}, veRL \cite{sheng2025hybridflow}, and AReaL \cite{fu2025areal}) only support single-agent scenarios, and even recent MARL frameworks represented by MARTI \cite{marti2025} are customized for specialized MARL algorithms and incremental extensions of single-agent RL. They naively optimize the rollout or training phase and lack end-to-end optimization considering the characteristics of MARL.
Besides, their implementation fails to support cross-node placement for individual agents, and LLM-based agents are confined to limited parameters (e.g., 3B), impeding large-scale deployment.

Through profiling of real-world MARL workloads, we identify three MARL-specific challenges that differ fundamentally from those of general agentic RL. 
\textit{(1) Topology-induced rollout skew}. Rollout request scheduling exhibits a severe imbalance among agents during parallel trajectory generation. Core agents appearing on the critical paths are repeatedly involved and incur request queuing, while auxiliary agents remain underutilized. 
\textit{(2) Critical path synchronization barriers}. MARL amplifies long-tail latency through multi-agent interaction trajectories. Policy training has to wait for the slowest path to complete, leaving expensive compute resources idle. 
\textit{(3) Dynamic training demand}. Different subsets of agents could be activated for policy training based on the collected experience. The computing and memory resources occupied by inactive agents are wasted under static allocation strategies, resulting in poor resource utilization.

Our key insight is that these challenges are not independent issues but stem from the graph-structured nature of MARL. The fundamental execution unit is not an isolated request or agent, but a dependency structure formed by agent invocations, outputs, and policy-versioned experiences \cite{xie2025ai}. 
Existing RL frameworks rely on uniform scheduling and static resource allocation during both rollout and training phases, which do not jointly optimize the above MARL-specific challenges based on the dynamic and collaborative characteristics of MARL.
This motivates us to design an intermediate representation that makes the multi-agent dependency structure explicit and uses it to guide rollout, orchestration, and training of MARL.

\enlargethispage{15pt}
To fill this gap, we propose an end-to-end MARL training framework, termed FlexMARL, that co-designs rollout, training, and their orchestration for large-scale LLM-based MARL. FlexMARL introduces the multi-agent execution graph (MEG) and materializes the graph representation as the \textit{MEG Store}.
Each vertex represents an agent invocation, and each edge represents a data dependency. 
During rollout, FlexMARL uses these dependencies to decide which downstream agents can be executed. During training, FlexMARL groups vertices by agent and policy version, and dispatches them for independent policy optimization.
The MEG Store provides decentralized multi-table storage and serves as the logical data hub for the following three components.

The \textit{Rollout Engine} employs dependency parallel sampling that enables concurrent trajectory generation as soon as upstream dependencies are satisfied.
To address topology-induced rollout skew, the Rollout Engine monitors multi-agent workloads and then adopts hierarchical load balancing to flexibly adjust the number of inference instances across agents.
The \textit{Joint Orchestrator} adopts the disaggregated architecture and decomposes the global batch at the granularity of agents. The agent-level asynchronous pipeline hides critical path rollout behind training through incremental gradient caching and unified updating, preserving multi-agent policy version consistency.
The \textit{Training Engine} exploits sparse training activation of MEG by collectively managing all training processes for an individual agent. Hardware resources are allocated to active agents only where and when needed, thus adapting to dynamic training demands and improving resource utilization.
We abstract multi-tier pathways into a location-agnostic \texttt{Set}/\texttt{Get} API, which is used to efficiently swap the training states of suspended and resumed agents.

During each step, the Rollout Engine generates multi-agent trajectories via parallel sampling, with inference instances scaling elastically. These trajectories are collected into the MEG Store, where the Joint Orchestrator dispatches micro batches asynchronously to the Training Engine. The Training Engine allocates resources on demand and swaps states for policy training. The updated models are synchronized back to inference instances to start the next step.
Overall, we make the following contributions in this paper:
\enlargethispage{10pt}
\begin{itemize}
\setlength{\itemsep}{1.5pt}
\setlength{\parsep}{1.5pt}
\setlength{\parskip}{1.5pt}
    \item We identify three MARL-specific challenges: topology-induced rollout skew, critical path synchronization barriers, and dynamic resource demand, which can be attributed to the graph-structured nature of MARL. Based on this insight, we introduce the MEG and materialize the graph representation as the MEG Store.
    \item We propose FlexMARL, the first framework to provide end-to-end optimization based on unique characteristics of large-scale LLM-based MARL.
    Built around the MEG Store, FlexMARL designs dependency parallel sampling with hierarchical load balancing for rollout, agent-level asynchronous pipeline for orchestration, and on-demand resource allocation for training. 
    \item We implement and deploy FlexMARL on a production cluster. Evaluations on public datasets and e-commerce multi-agent assistants show that FlexMARL provides up to 7.3$\times$ speedup and improves hardware utilization by up to 5.6$\times$ compared to existing MARL frameworks.
\end{itemize}

%% file: content/background.tex
\subsection{Multi-Agent Reinforcement Learning}
Traditional non-LLM MARL is largely restricted to well-structured environments with predefined action sets \cite{martinez2014marl,canese2021multi,zhang2021multi}. In contrast, LLM-based MARL represents a paradigm shift from low-level control optimization to knowledge-driven, high-level collaboration \cite{yan2026data}. The agents are instantiated as LLMs with inherent natural language understanding and reasoning capabilities. These LLM agents process unstructured information and generate token sequences as actions. As a result, LLM-based MARL effectively decomposes task complexity, enabling agents to learn collaboratively through interactions with other agents \cite{zhu2025lamarl,liu2025llm}. Recent advancements have highlighted its effectiveness in solving long-horizon tasks such as code generation \cite{zhao2025mage,pan2025codecor}, mathematical reasoning \cite{zhang2025debate4math,li2026dynadebate}, and network management \cite{wu2024netllm,wang2025intent}, which are challenging for conventional neural network policies.

\begin{figure}[t]
\centering
\subfigure[Queued Request]{
\centering
\includegraphics[width=0.47\linewidth]{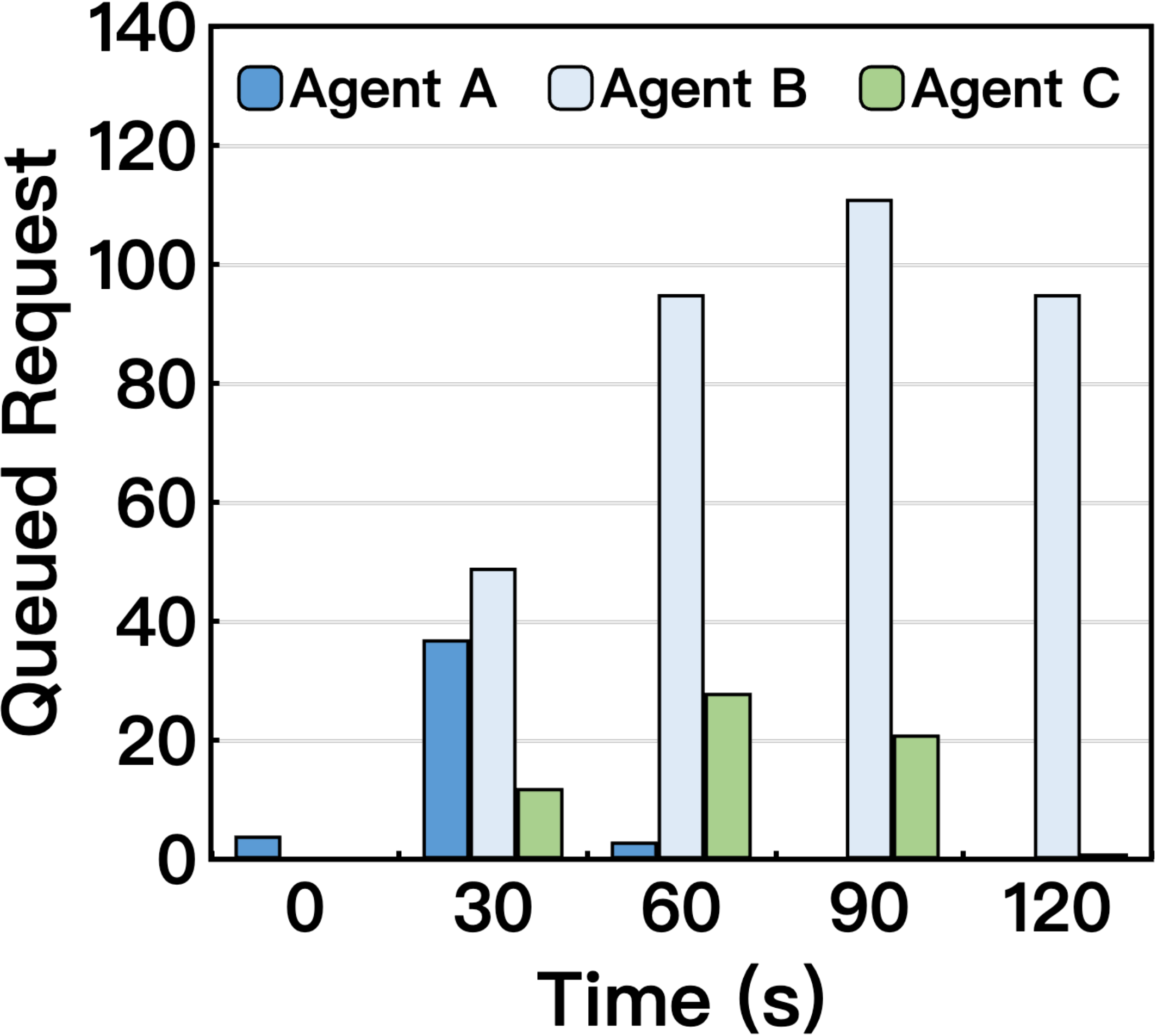}
}
\subfigure[Query Response Time]{
\centering
\includegraphics[width=0.475\linewidth]{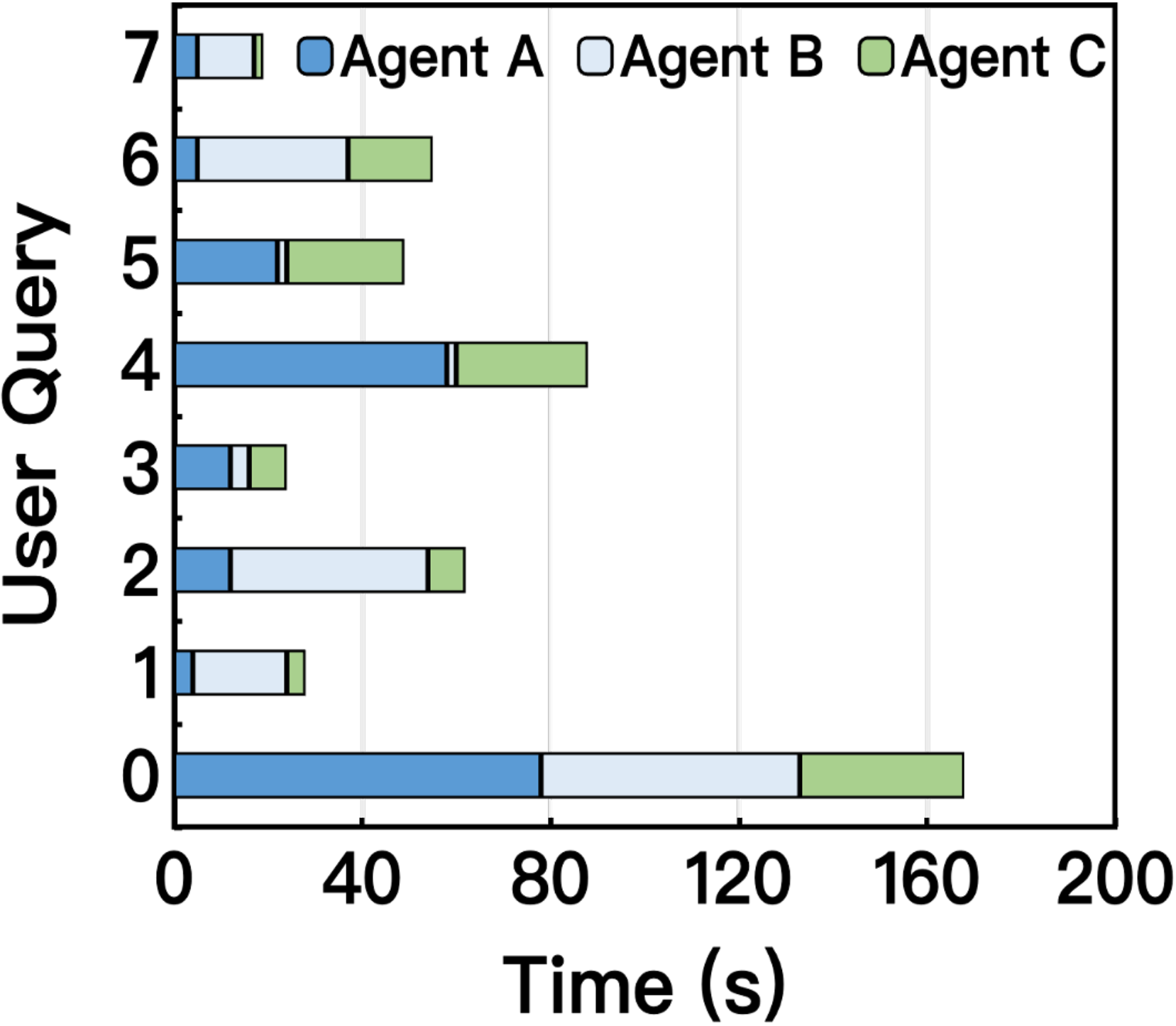}
}
\vspace{-4mm}
\caption{Preliminary experiments demonstrating the MARL characteristics.}\label{fig:pre_experiment}
\vspace{-10pt}
\end{figure}

\begin{table*}[!t]
\centering
\renewcommand\arraystretch{1.1}
\caption{Comprehensive comparison: FlexMARL vs. Existing RL/MARL frameworks.}
\label{tab:system_comparison}
\resizebox{0.99\textwidth}{!}{
\begin{tabular}{c|ccc|ccc}
\toprule
\multirow{2}{*}{\textbf{Frameworks}}   & \multicolumn{3}{c|}{\textbf{Basic Features}}                                                                                                                                                               & \multicolumn{3}{c}{\textbf{End-to-End Optimization}}                                                                                                                                                                        \\ \cline{2-7} 
                              & \textbf{\begin{tabular}[c]{@{}c@{}}Multi-Agent\\ Support\end{tabular}}& \textbf{\begin{tabular}[c]{@{}c@{}}Algorithm\\ Decoupling\end{tabular}} & \textbf{\begin{tabular}[c]{@{}c@{}}Large-Scale\\ Deployment\end{tabular}} & \textbf{\begin{tabular}[c]{@{}c@{}}Topology-Induced \\Rollout Skew\end{tabular}} & \textbf{\begin{tabular}[c]{@{}c@{}}Critical Path \\Sync Barriers\end{tabular}} & \textbf{\begin{tabular}[c]{@{}c@{}}Dynamic\\ Training
Demand \end{tabular}} \\ \hline
\textbf{\begin{tabular}[c]{@{}c@{}}General-Purpose RL Frameworks\\ (e.g., OpenRLHF \cite{hu2024openrlhf}, veRL \cite{sheng2025hybridflow}, AReaL~\cite{fu2025areal})\end{tabular}}  & \ding{55}                                                              & \ding{51}                                                               & \ding{51}                                                                 & \ding{55}                                                                &  \ding{51}                                                                      & \ding{55}                                                                        \\ \hline
\textbf{\begin{tabular}[c]{@{}c@{}}Specialized MARL Frameworks\\ (e.g., MARTI \cite{marti2025})\end{tabular}}   & \ding{51}                                                              & \ding{55}                                                               & \ding{55}                                                                 & \ding{55}                                                                & \ding{55}                                                                       & \ding{55}                                                                        \\ \hline
\rowcolor{green!20} \textbf{FlexMARL (Ours)}               & \ding{51} & \ding{51} & \ding{51} & \ding{51} & \ding{51} & \ding{51}                                        \\ 
\bottomrule
\end{tabular}}
\vspace{2mm}
\end{table*}

The typical workflow of LLM-based MARL contains repeated cycles across rollout and training phases.
In the rollout phase, each LLM agent generates responses based on its current policy parameters, role-specific prompts, and environmental observations \cite{zhao2025stronger}. Multi-agent interactions produce the trajectories that capture their reasoning processes. These experiences are then evaluated using a rule-based or learned reward model that reflects task correctness, collaboration effectiveness, and alignment with global objectives. After credit assignment across agents, the collected experiences are organized into batches and dispatched to corresponding agent policies for training \cite{hu2024measuring}. Each agent independently optimizes its policy model using RL algorithms (e.g., GRPO \cite{shao2024deepseekmath}) to maximize the expected cumulative reward. The updated models are synchronized via device-to-device (D2D) communication for the next step of rollouts.

\subsection{Observations of MARL Characteristics}
Despite the advancements in MARL algorithms, the underlying infrastructure bottlenecks hinder large-scale training efficiency. 
To reflect the characteristics of LLM-based MARL, we conduct extensive profiling of real-world multi-agent merchant assistant workloads. The setup of preliminary experiments is similar to that in Section \ref{sec:Evaluation}.
Three observations are as follows.

\textbf{Observation \#1: MARL exhibits imbalanced rollout load distributions across agents.}
A salient characteristic of MARL is that agents are typically assigned different roles due to task-specific collaboration mechanisms. During the experience collection phase, MARL needs to handle a large number of continuous rollout requests, exhibiting different interaction patterns based on functional roles. 
The rollout load among distinct agents shows a highly skewed distribution. 
We select three representative agents and track the number of their queued requests over time.
As illustrated in Figure \ref{fig:pre_experiment}(a), a small portion of core agents are repeatedly invoked, handling over 76\% of the total rollout requests, whereas other auxiliary agents process less than 24\% of the requests. Such an imbalanced workload causes the core agents' requests to continuously queue, which prolongs the trajectory sampling time and becomes a bottleneck.

\textbf{Observation \#2: MARL suffers from long-tail multi-agent interaction latency.} 
Due to differences in task complexity and agent decision paths, multi-agent interaction latency exhibits a pronounced long-tail distribution \cite{li2026qurl}.
As shown in Figure \ref{fig:pre_experiment}(b), most multi-agent interactions are completed in a short time, but the longest response time of user queries can reach nearly 170s, which dominates the overall experience collection time. Under synchronous training paradigms, policy updating is blocked by ``stragglers'' due to strict serial dependency. Policy training is triggered until all trajectories are collected, causing rollout-training synchronization barriers.

\textbf{Observation \#3: MARL demonstrates dynamic resource demands during policy training phases.}
The policy training phase may activate different subsets of agents for gradient computation based on the collected samples. 
If we allocate dedicated resources to each agent using a fixed mapping strategy, the computing and memory resources occupied by inactive agents are wasted rather than reallocated to active agents with high current demand.
Our preliminary experiments also confirm that the static allocation strategy lacks a global perspective, resulting in average hardware utilization of only 18.8\% during policy training.

\subsection{Key Insights}
The above three observations reveal that large-scale LLM-based MARL introduces new system challenges that cannot be addressed by naively extending single-agent RL infrastructure. 
We realize that the fundamental execution unit of MARL is not an isolated request or an isolated agent, but a set of agent invocations (vertices) connected by data dependencies (edges). 
This graph dependency structure exposes and amplifies the observed challenges.

\textbf{Revisiting Observation \#1: Rollout load imbalance is induced by graph topology.} 
Load is unevenly distributed across different graph vertices. Core agents correspond to high-degree vertices that appear on the critical paths of many trajectories, while auxiliary agents correspond to low-degree vertices invoked by only a few paths. Request-based scheduling cannot resolve inherent skew caused by graph topology.

\textbf{Revisiting Observation \#2: Synchronization barriers stem from graph critical paths.} 
The long-tail distribution reported in Figure~\ref{fig:pre_experiment}(b) is explained as the dispersion of path lengths across the graph.
The overall rollout time of multi-agent interactions is determined by the longest path in the graph (critical path), not by individual agent invocations. 
Under batch-level synchronization paradigms, policy optimization has to wait for the slowest dependency path to complete, inducing severe pipeline bubbles.

\textbf{Revisiting Observation \#3: Low resource utilization arises from sparse subgraph activation.} 
Only subgraphs corresponding to agents with sufficient accumulated experience are activated for policy optimization. The active vertex groups are sparse and change per training step. 
Static allocation strategy wastes resources on agents that are present in MARL workloads but inactive in the current subgraph.

\subsection{Limitations of Existing Works}
Existing RL frameworks generally fall into two categories, neither of which performs end-to-end optimization based on the graph structure nature of MARL.
On the one hand, general-purpose RL frameworks (such as OpenRLHF~\cite{hu2024openrlhf}, AReaL~\cite{fu2025areal}, and veRL~\cite{sheng2025hybridflow}) are primarily designed for a single model and cannot capture cross-agent dependencies, agent load skew, or dynamically activated training subsets in LLM-based MARL.
On the other hand, specialized MARL frameworks \cite{zhao2025stronger,feng2026dr,marti2025} only provide basic multi-agent training capabilities but treat the underlying infrastructure as ad-hoc extensions of single-agent RL, without end-to-end and MARL-specific optimization.
As a result, they rely on uniform scheduling and static resource allocation during both rollout and training phases.
Such designs suffer from load imbalance when rollout workloads are skewed across heterogeneous agents, induce synchronization barriers when long-tail dependency paths prolong batch completion, and waste compute/memory resources when only a subset of agents accumulate sufficient training samples.
Furthermore, their implementations lack cross-node deployment support and are limited to small-scale models. For instance, MARTI \cite{marti2025} is deployed on a 3-node cluster, and the largest model participating in MARL is Qwen2.5-3B. Such constrained configurations prevent the exploration of performance gains from larger agent ensembles and more powerful models, thereby limiting the scalability of MARL.

As summarized in Table~\ref{tab:system_comparison}, existing RL frameworks fail to simultaneously address the challenges exhibited by large-scale MARL training, including topology-induced rollout skew (Observation \#1), critical path synchronization barriers (Observation \#2), and dynamic training demand (Observation \#3). This reveals the gaps between algorithmic requirements and existing infrastructure support.

%% file: content/overview.tex
\begin{figure}[t]
    \centering 
    \includegraphics[width=1\linewidth]{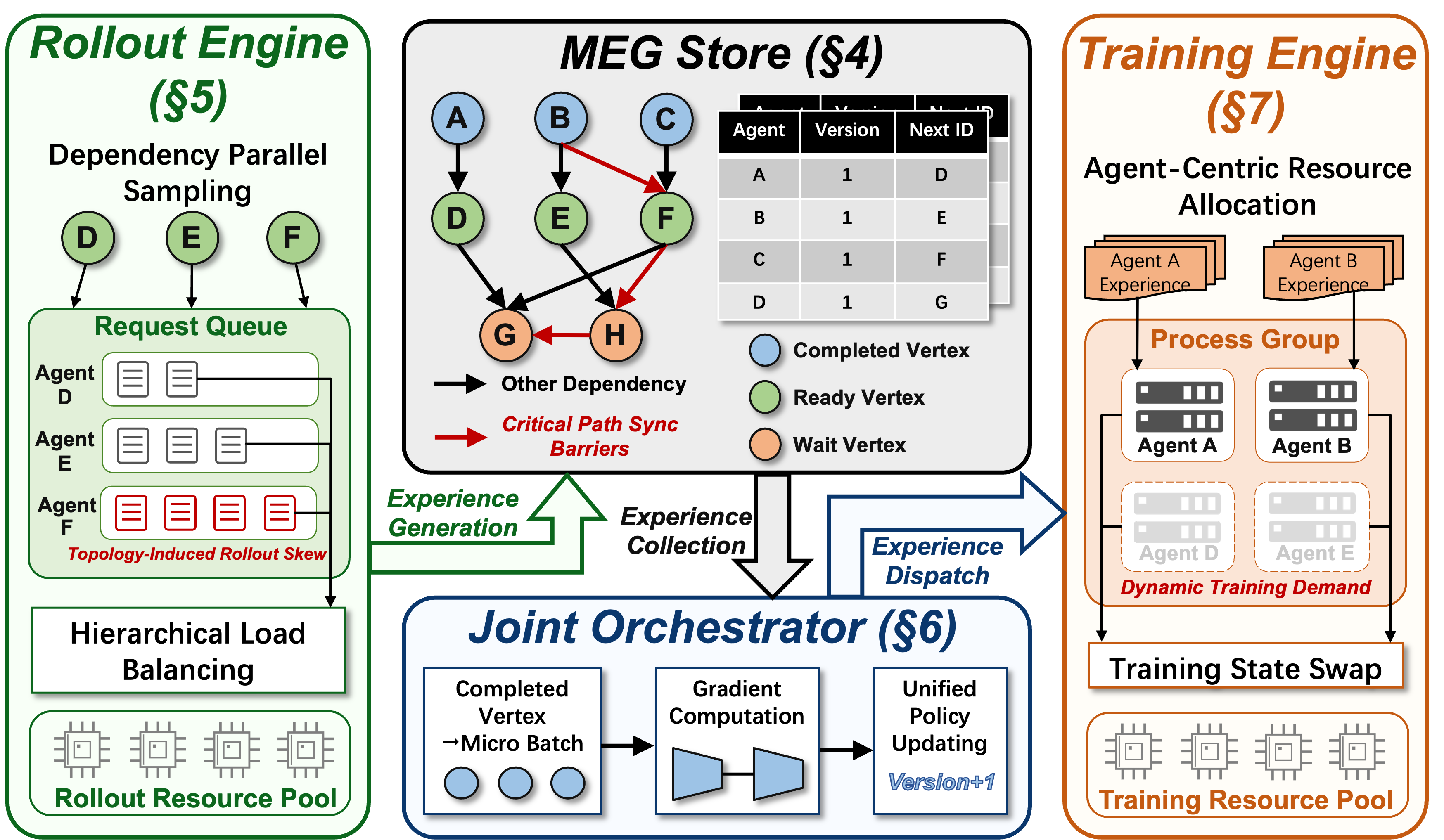}
    \vspace{-4mm}
    \caption{Overview of FlexMARL.}
    \label{fig:overview}
\end{figure}

Motivated by the graph-structured nature of MARL, we propose FlexMARL, a graph-driven rollout-training co-designed framework that provides end-to-end optimization for large-scale LLM-based MARL. 
As illustrated in Figure \ref{fig:overview}, FlexMARL consists of four tightly integrated components:

$\bullet$ \textbf{MEG Store (\S\ref{sec:MEG}).}  
FlexMARL uses the MEG as an intermediate representation and materializes the graph in the MEG Store, which organizes agent invocations, data dependencies, policy versions, and training status.
As the concrete runtime materialization, the MEG Store employs decentralized multi-table storage and serves as the data hub connecting system components by supporting incremental experience generation during rollout and efficient retrieval during training.

$\bullet$ \textbf{Rollout Engine (\S\ref{sec:Load-Aware Rollout Engine}).} 
To improve inference efficiency, our engine employs a dependency parallel sampling scheme that schedules rollout requests according to satisfied upstream dependencies, allowing different trajectory branches to progress concurrently.
Based on request queues, hierarchical load balancing elastically adjusts the number of intra-agent/inter-agent inference instances via D2D transmission, thus alleviating topology-induced rollout skew.

$\bullet$ \textbf{Joint Orchestrator (\S\ref{sec:Joint Orchestrator}).} 
The orchestrator adopts the disaggregated architecture to allocate dedicated resource pools and overlaps rollout and training at the agent granularity. 
Instead of blocking training until the rollout of MEG completes, it decouples gradient computation from policy update. Policy training is triggered incrementally for each agent as soon as part of the experience is available in the MEG Store, thereby eliminating the synchronization barriers caused by the long-tail critical path.

$\bullet$ \textbf{Training Engine (\S\ref{sec:Resource-Efficient Training Engine}).} Since the agents’ experiences become ready at non-deterministic times, different subgraphs of MEG are activated during training. 
FlexMARL exploits this sparsity by agent-centric resource allocation.
When an agent requires policy optimization, its process group is dynamically scheduled onto available resources. 
We further implement efficient state swap mechanisms that transfer training states (e.g., weights and optimizer states) between device and host memory, adapting to large-scale scenarios where a large number of agents need to be updated.

The collaboration process of the four components is described as follows. 
The Rollout Engine continuously retrieves agent invocation tasks whose all upstream dependencies have been satisfied, dispatches these tasks to elastic inference instances, and writes back the output responses to the MEG Store.
This process triggers new downstream agents incrementally until the complete trajectory is generated. The agent-level experiences are also collected in the MEG Store with version tracking. The Joint Orchestrator dispatches micro batches of experiences to the Training Engine with ongoing rollouts.
Based on available cluster capacity, the Training Engine concurrently allocates hardware resources for active agents to perform policy optimization. After processing a full batch, accumulated gradients are aggregated for unified parameter updates. Finally, new policy models are synchronized to inference instances via D2D interconnects.

%% file: content/graph_abstraction.tex
\subsection{Graph Representation}
Existing RL systems typically organize experience as flat trajectories or batches.
However, LLM-based MARL differs from single-agent RL in a structural way. MARL contains multiple role-specialized agents whose invocations depend on outputs from other agents, and some branches may proceed in parallel. 
A flat trajectory representation makes it difficult to exploit the topology-induced rollout skew, critical path synchronization barriers, and dynamic training demand. 
To this end, FlexMARL models MARL workloads using the MEG, which is a directed graph $(\mathcal{N}, \mathcal{E})$. Each vertex $n\in\mathcal{N}$ is one agent invocation with specific policy version $V_n$, input prompt $I_n$, output response $O_n$, and reward $R_n$.
Each edge $e_{n,m}\in\mathcal{E}$ represents a data dependency, meaning that the downstream agent $m$ can be executed only after the upstream agent $n$ has produced the response.  
The directed property follows the execution semantics of MARL workloads. An agent invocation consumes previously generated context and produces a new output for downstream agents. MEG handles learning feedback across training steps through updated policy versions.

The MEG supports two views over the same set of vertices.
In the \textit{rollout view}, vertices execute in topological order, and edges enforce data dependencies between agents. They determine when an agent becomes ready and which downstream agents can be invoked. 
Different vertices have different out-degrees and in-degrees, indicating a skewed rollout load.
The critical path dominates the rollout latency of multi-agent interactions.
In the \textit{training view}, the completed vertices are grouped by agent identifier and policy version. The edges no longer impose execution dependencies because policy optimization is performed independently for each agent after experience collection and reward assignment. Training activation is sparse and dynamic since only agents that have sufficient experience perform policy optimization.

\begin{figure}[t]
    \centering 
    \includegraphics[width=1\linewidth]{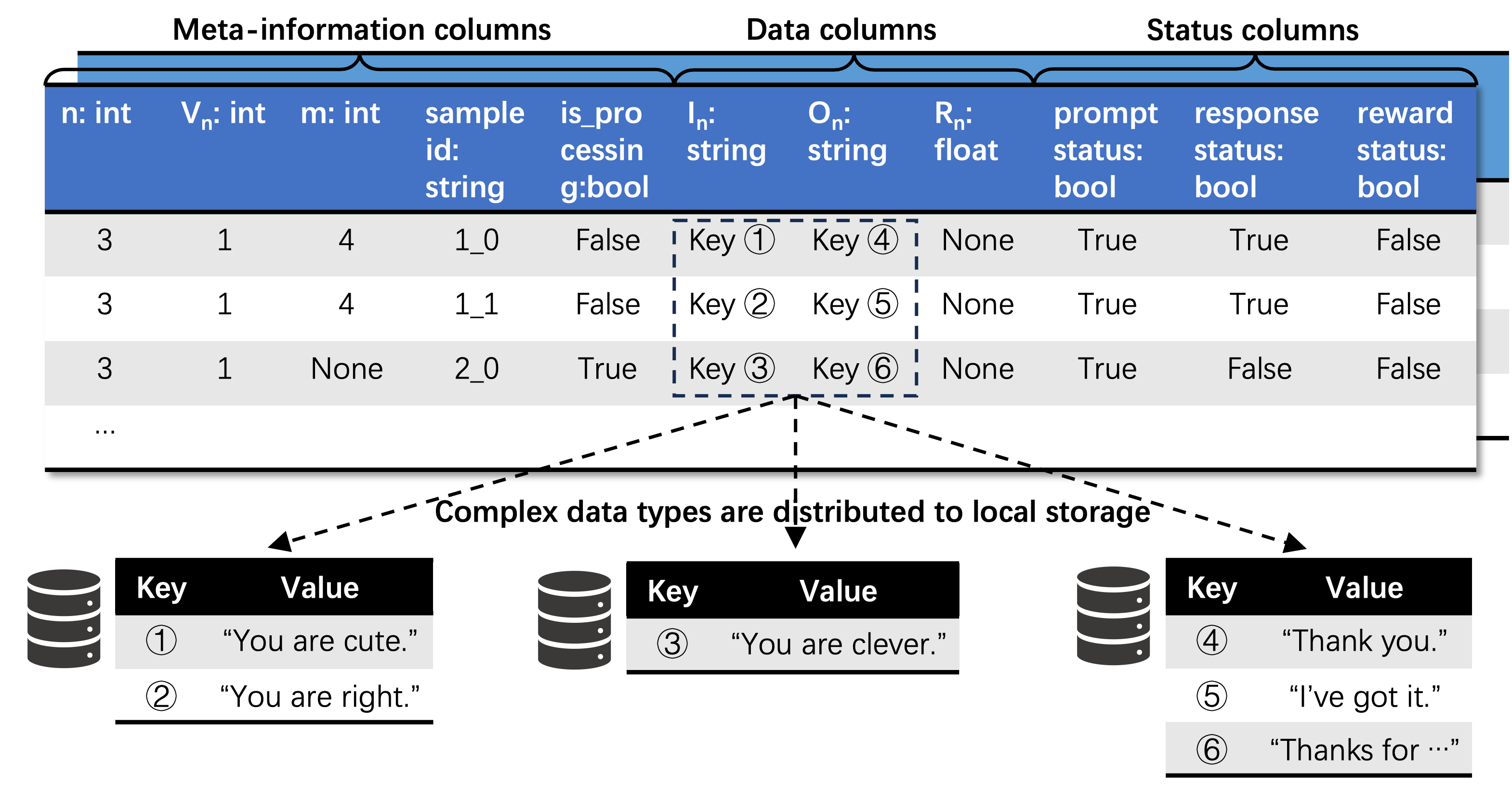}
    \vspace{-5mm}
    \caption{Multi-table organization of the MEG Store.}
    \label{fig:SampleTable}
     \vspace{-10pt}
\end{figure}

\subsection{Multi-Table Materialization}
FlexMARL materializes the graph representation through the MEG Store, which supports MARL’s requirements for data isolation, cross-agent dependency, and policy version tracking.
The MEG Store is a logical data hub rather than a centralized service.
As illustrated in Figure~\ref{fig:SampleTable}, the MEG Store is organized by decentralized multi-table storage at the granularity of graph vertices $n\in\mathcal{N}$.
Therefore, increasing the number of agents does not route all experience payloads through one central bottleneck.
In a decentralized table, each row contains three categories of columns: meta-information columns, data columns, and status columns. 
Specifically, meta-information columns store essential metadata for each sample, including the agent identifier $n$, policy version $V_n$, next agent identifier $m$, sample identifier, and a processing flag indicating whether a sample has been read but not yet updated. 
The next agent identifier $m$ represents the dependency edge $e_{n,m}\in\mathcal{E}$ in MEG.
The sample identifier ensures global uniqueness and traceability of the form ``$\texttt{sample\_id} = \texttt{\{input\_id\}\_\{trajectory\_id\}}$'', where \texttt{input\_id} denotes a globally incrementing identifier for each sample and \texttt{trajectory\_id} is the index of the response within a trajectory. 
Data columns contain user-defined fields that store sample content, such as input prompt $I_n$, output response $O_n$, and reward $R_n$. Each data column is paired with a boolean status column that specifies whether its corresponding value has been fully generated.

To support lightweight querying while accommodating extensible user-defined fields across MARL algorithms, the MEG Store adopts type-aware hybrid storage that selects value-based or reference-based storage based on data type. All meta-information and status fields use simple data types (including int, float, and bool), which are stored by value for fast in-table access.
Complex data types (including strings, lists, and tensors) are stored by reference, and the table only records the location key.
FlexMARL fetches the keys from the table and then retrieves the corresponding values directly.

%% file: content/rollout_engine.tex
\subsection{Dependency Parallel Sampling}

The processing of rollout requests follows a sequential execution model.
Each agent may generate additional candidate samples for the same prompt. These samples are then propagated through downstream agents to generate complete multi-agent trajectories. The next trajectory can be generated only after the entire rollout of the current trajectory has finished \cite{zhang2026train}. While this design preserves clear execution semantics, it severely limits hardware utilization and system throughput, especially in large-scale MARL settings.

To eliminate these inefficiencies, our Rollout Engine introduces dependency parallel sampling.
Once the upstream dependencies of a rollout request have been satisfied, it can be dispatched to an inference instance of downstream agents immediately. 
Other queries or branches are independent of the completion state of the current query. 
In other words, multiple queries and/or multiple trajectory generation for the same query are allowed to execute rollout in parallel.
After the agent $n$ generates a response, the Rollout Engine writes the output $O_n$ and the corresponding status fields back to the MEG Store. The downstream agents are notified through the next agent identifier $m$. The MEG Store is used not only to collect the final trajectories, but also to mediate intermediate data movement between agents. 
This procedure continues until complete multi-agent interaction trajectories are generated.
Dependency parallel sampling converts the experience generation from a query-level blocking model into a fully parallelized scheduling model.  
All rollout requests compete dynamically for inference resources subject to data dependencies without unnecessary waiting, thereby significantly improving rollout efficiency.

\begin{figure}[t]
    \centering 
    \includegraphics[width=1\linewidth]{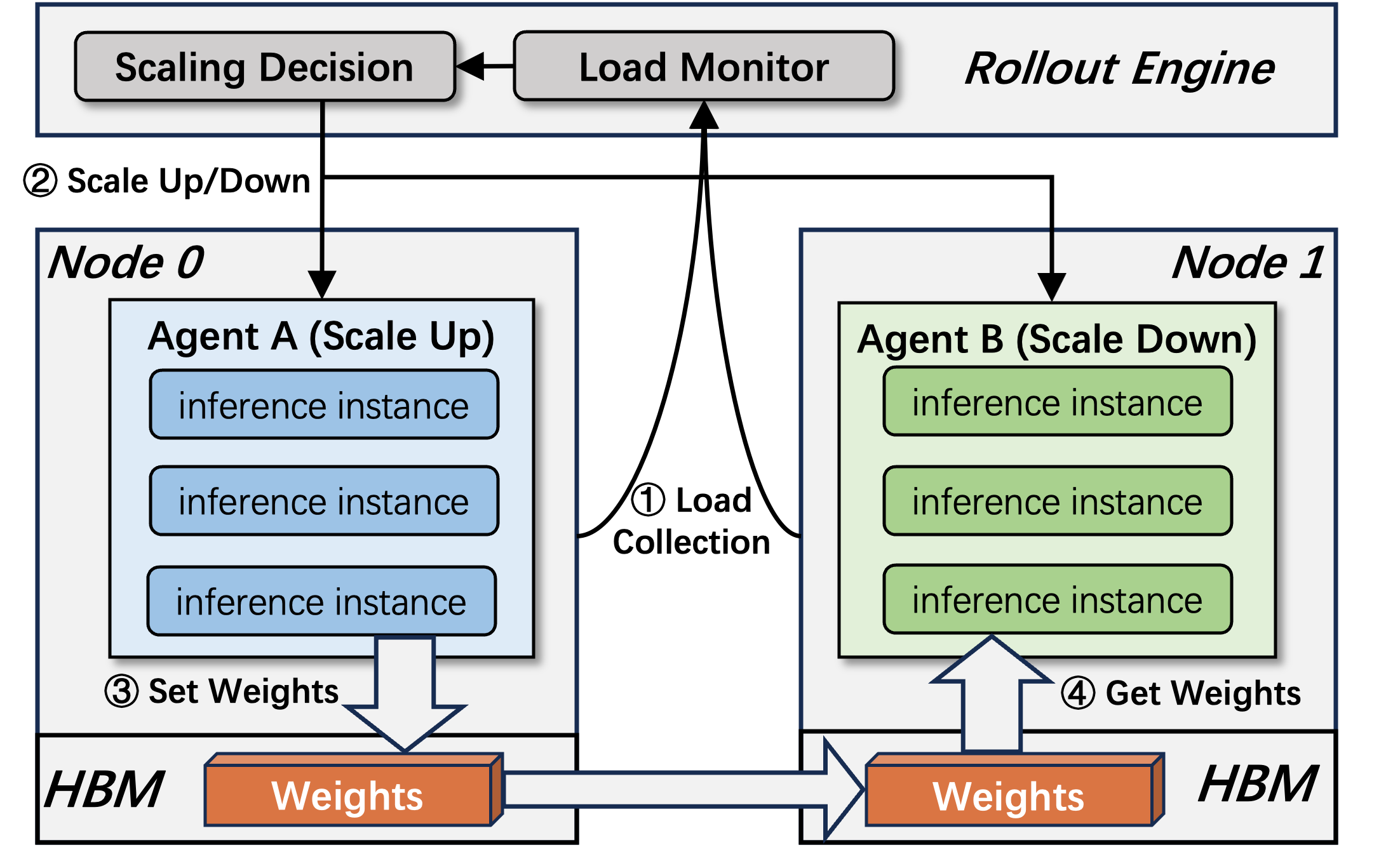}
    \caption{Inter-agent load balancing in the Rollout Engine.}
    \label{fig:scaling}
    \vspace{-1mm}
\end{figure}

\subsection{Hierarchical Load Balancing}
Dependency parallel sampling increases available query parallelism, but it also makes topology-induced load skew more visible. 
Due to agents' functional specialization, core agents may appear on many dependency paths and exhibit large in-degree/out-degree in the MEG. Auxiliary agents are rarely invoked, resulting in persistent queuing delays at overloaded agents and underutilization at others \cite{xie2025ai}.
To mitigate load imbalance, Rollout Engine provides a two-tier elastic scaling that balances rollout load both within and across agents.

\textbf{Intra-Agent Load Balancing.}
For each agent, we deploy multiple inference instances to handle its concurrent rollout requests. The Rollout Engine employs a min-heap data structure of request queues to track the instantaneous load of backend inference instances.
A new rollout request will be dispatched to the instance with the lowest current load.
This greedy scheduling scheme achieves balanced request distribution within inference instances of individual agents.
Moreover, the Rollout Engine ensures fault tolerance by removing completed requests, canceling timed-out ones, and re-queuing unfinished ones.

\textbf{Inter-Agent Load Balancing.}
While intra-agent balancing mitigates contention within a single agent, it cannot resolve resource starvation across agents with heterogeneous demands.
As shown in Figure \ref{fig:scaling}, the Rollout Engine polls the request queue lengths across all agents. 
We analyze these load metrics and then compute the load disparity between the highest and lowest load agents. When this disparity exceeds a configurable threshold $\Delta$, a scaling operation is triggered to migrate inference capacity from underutilized agents to overloaded ones \cite{hu2025flb}. Specifically, the number of migrated instances equals the difference in queue lengths, subject to the constraint that each agent retains at least one active inference instance to preserve liveness. 
The conservative strategy prevents transient load oscillation while ensuring service availability for all agents.
Migrating inference capacity requires rapid model weight transfer to newly allocated instances.  
FlexMARL abstracts multi-tier pathways into a unified and location-agnostic \texttt{Set}/\texttt{Get} API, which hides the complexities of D2D, H2D, and D2H communication. 
When a scaling operation is triggered, the agents requiring scaling up (Agent A) publish model weights via \texttt{Set} API, while the reallocated instances of scale-down agents (Agent B) overwrite their original weights via \texttt{Get} API. 
The implementation of \texttt{Set}/\texttt{Get} API will be detailed in \S\ref{sec:implementation}.
Finally, the migrated instance detaches from the donor Agent B and registers with the target Agent A to accept new rollout requests.

%% file: content/orchestrator.tex
Existing training frameworks for LLM-based MARL typically employ the colocated architecture, where the rollout and training phases share the identical resource pool. 
In fact, the rollout and training phases have different task characteristics. The rollout phase is memory-bound, while the performance of the training phase is limited by computational power. 
It is suboptimal to allocate the same resources to two distinct phases, leading to resource underutilization.
Beyond resource conflicts, the colocated architecture also degrades MARL's training efficiency because the rollout and training phases alternate for time-division multiplexing of resources. The shared devices cannot transition to the training state due to critical path synchronization barriers \cite{wang2025reinforcement,xiaomi2025mimo,gao2025rollpacker,zhu2025megascale}. This problem is exacerbated in MARL compared to single-agent RL, as multi-agent interactions amplify long-tail latency.
In light of the above discussion, the Joint Orchestrator advocates the disaggregated architecture for LLM-based MARL, which allocates dedicated, physically separate resource pools for the rollout and training phases. This decoupling design avoids resource conflicts and allows independent scaling according to phase characteristics.

To fully unlock the potential of disaggregation, the orchestrator needs to determine the collaborative paradigm to overlap the rollout and training phases.
Although asynchronous pipelines have been widely adopted in single-agent RL \cite{zhong2025streamrl,han2025asyncflow,fu2025areal}, they operate on a single policy model and flat trajectories. Applying them directly to MARL introduces two limitations.
Firstly, these schemes require the entire multi-agent interactions to complete before any experience can be used for training.
Secondly, single-agent asynchronous pipelines only maintain consistency for a single policy, while MARL training involves multiple policy models. 
In fact, MARL is more sensitive to policy staleness compared with single-agent RL because the bias propagates along the dependency paths rather than remaining local to one agent invocation \cite{gao2025rollpacker}. 
If different vertices of the MEG observe incompatible policy versions, the collected experience no longer corresponds to consistent multi-agent interaction, ultimately causing significant accuracy degradation.

To this end, the Joint Orchestrator adopts an agent-level asynchronous pipeline, which decomposes the global batch at the granularity of agents.
Concretely, completed vertices become visible to the training view of MEG as soon as part of the experience is available. 
Thanks to the multi-table organization, once the agent-level accumulated experience in the MEG Store exceeds a predefined micro batch size, the Joint Orchestrator dispatches them to the corresponding training process for independent gradient computation on heterogeneous policy models. 
In this way, policy training is triggered incrementally in a fine-grained manner without waiting for the entire batch \cite{chen2025crosspipe}.
This allows training to start much earlier and execute in parallel with the remaining rollout requests, thereby alleviating pipeline bubbles \cite{feng2025optimus} and synchronization barriers caused by long-tail critical path.

More importantly, we carefully manage policy versions by decoupling \emph{when gradients are computed} from \emph{when policy updates are committed}.
Each agent computes its gradients in advance, but will not immediately update its policy models, which creates an implicit frozen window for parallelizing rollout and training without introducing multi-agent policy version inconsistency.
Policy model parameters remain frozen during all micro batch gradient computation, and there is no additional policy version.
Therefore, different vertices leverage consistent and fresh policy models for the ongoing rollouts, avoiding the cross-agent staleness issue.
After processing a full batch, all agents aggregate the cached gradients to perform unified policy updating, and policy versions are also updated in the MEG store. 
The updated parameters are synchronized to all inference instances via high-bandwidth D2D interconnects, ensuring that the next rollout also uses the up-to-date policy model.

%% file: content/training_engine.tex
\subsection{Agent-Centric Resource Allocation}
In the training view of MEG, completed vertices are grouped by agent and policy version, and only agents with sufficient experience in the MEG Store need to be activated. 
As a result, the set of agents undergoing policy optimization and their entry is non-deterministic.
Static allocation inevitably leads to resource waste, as inactive agents monopolize memory and computing resources throughout training.
To accommodate the sparse subgraph activation pattern, our Training Engine integrates agent-centric allocation to bind training resources only where and when needed.
In particular, we implement time-division multiplexing for training resources via process group abstraction, which encapsulates all training processes associated with an individual agent.
The Training Engine initializes a process group for each agent and
obtains handles for all required processes.
The training processes are activated, suspended, and resumed using a gang-scheduling strategy \cite{feitelson1997improved}, enabling collective lifecycle management.

Once the MEG Store accumulates sufficient training samples for a specific agent, the corresponding process group activates the training processes and dynamically schedules them to available resources within the cluster.
After an agent completes training for the current micro batch or when no new experiences are available, the process group is terminated. Unlike naive suspension, which retains process contexts in memory, our Training Engine terminates processes and immediately releases all associated hardware resources (including computing cores and HBM) back to the cluster pool.
When the Joint Orchestrator detects new experiences for the agent, the process group is re-created on available resources, and policy training resumes from the last checkpoint.
This ``suspend-to-destroy'' design ensures that storage and computation resources are never locked by inactive agents.
In this way, the training resource allocation in FlexMARL is aligned with agent-specific workloads rather than being preassigned to fixed hardware, maintaining high training throughput even with a massive number of potential agents.

\begin{figure}[t]
    \centering 
    \includegraphics[width=1\linewidth]{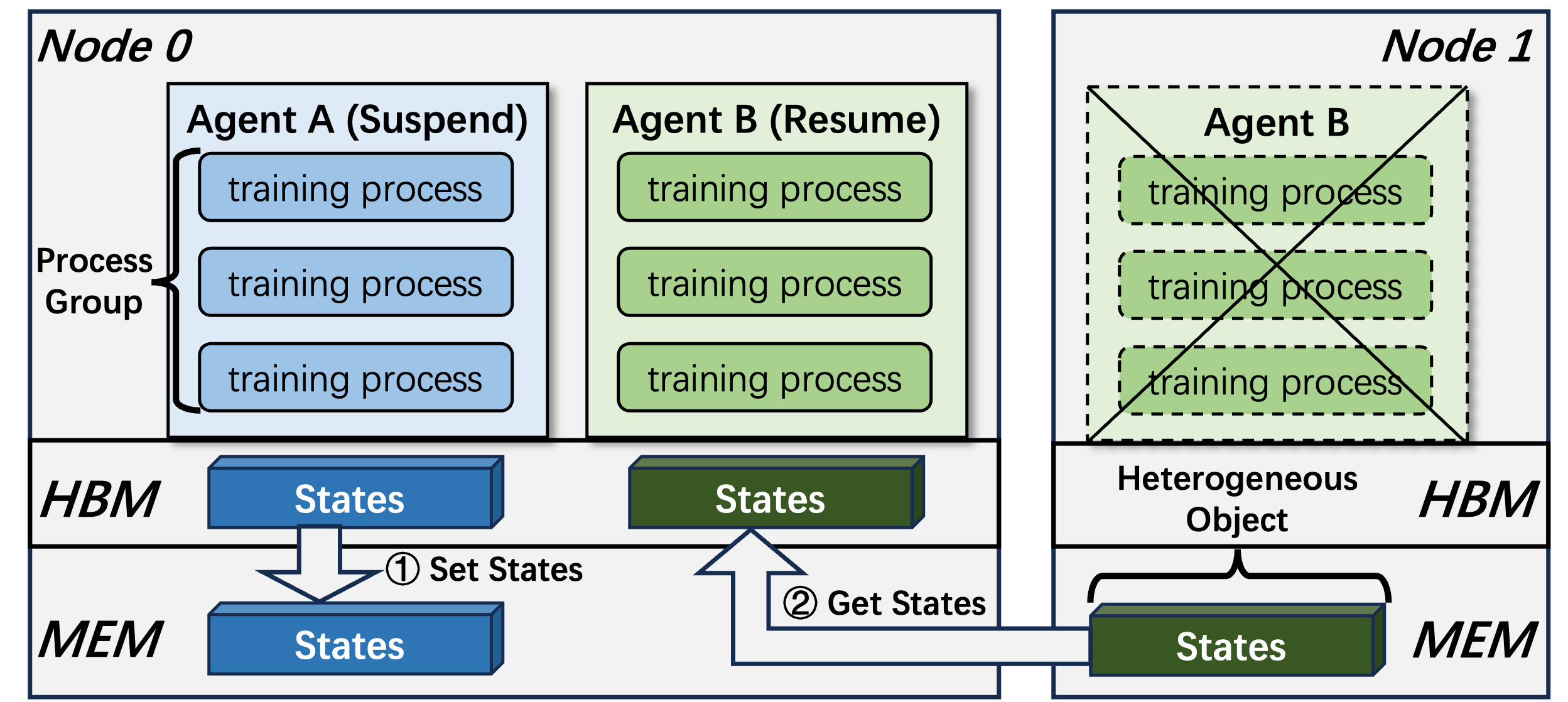}
    \caption{Efficient swap-in/out of training states for agent-centric resource allocation.}
    \label{fig:binding}
\end{figure}

\subsection{Training State Swap}
With dynamic reallocation of hardware resources, training states (including weights and optimizer states) cannot remain resident in HBM after process group suspension. Retaining the states for inactive agents would quickly exhaust device memory capacity and cause out-of-memory (OOM) failures, especially when the number of potential agents far exceeds available hardware resources. Consequently, on-demand resource binding necessitates efficient state swap between device and host memory.
Herein, we again leverage \texttt{Set}/\texttt{Get} API (\S\ref{sec:implementation}) to decouple logical state references from physical storage, enabling the Training Engine to track and access agent states consistently across distributed nodes.

As presented in Figure \ref{fig:binding}, the process group for the suspended agent (Agent A) triggers a checkpoint. Upon a \texttt{Set} operation, the current training states are offloaded from device memory to host memory as heterogeneous objects.
For the resumed agent (Agent B), the Training Engine exploits locality-aware deployment to minimize state migration latency.  
The process group is prioritized for scheduling to its previously deployed node. 
The Training Engine restores the states into the destination device memory through the \texttt{Get} API and creates the corresponding process group.
This training state swap process can be regarded as a trade-off between transfer overhead and resource utilization. Without state swap (transfer overhead=0), inactive agents would cause persistent waste. FlexMARL instead trades state swap overhead for improved resource utilization. 
However, for high-churn scenarios, FlexMARL tends to keep training states (including weights and optimizer states) of frequently activated agents resident in HBM and limit each device to serving a bounded number of agents, rather than repeatedly binding agents on a small set of devices. This distributes state swap operations, preventing transfer overhead from becoming a bottleneck.

%% file: content/implementation.tex
We implement FlexMARL in 8k lines of C/C++ code and 12k lines of Python code.
vLLM v0.9.1 is integrated as the rollout backend to ensure low-latency inference. Policy training is built on DeepSpeed v0.16.9 with ZeRO-3 enabled \cite{rajbhandari2020zero}.
To relieve developers from the complexities of function management and data transmission, FlexMARL implements the sleep/wakeup feature and heterogeneous object service based on the open-source openYuanrong \cite{openYuanrong}.
We deploy FlexMARL on a production 48-node cluster. Each node is equipped with 16 Ascend NPUs (64GB memory) and interconnected via HCCS \cite{zhou2025accelerating}.
At the foundational layer, the vendor-specific SDK and hardware abstraction library provide the necessary compute kernels, interfaced through a PyTorch adapter.

\subsection{Location-Agnostic \texttt{Set}/\texttt{Get} API}
Each heterogeneous object is accessed through key-value (KV) semantics, where the key provides a stable logical reference and the runtime resolves the corresponding physical location. The host and device memory are \textit{logically} unified to facilitate seamless data transfer across storage hierarchies. 
Each node in the cluster maintains a dedicated resident daemon, which is responsible for managing the distributed metadata of heterogeneous objects.
Based on this metadata service, hierarchical load balancing and training state swap leverage a \texttt{Set}/\texttt{Get} API for intra-domain communication (D2D) and cross-tier communication (H2D/D2H).

D2D transfer is used when rollout capacity is migrated across different agents, where newly allocated inference instances must rapidly load the scale-up agent's model weights.
The resident daemon utilizes a publish-subscribe coordination pattern where the invocation of the \texttt{Set} API triggers the registration of the device memory segment. This process records critical location metadata, including physical device address, memory offset, and node-level identifiers.
Data retrieval via the \texttt{Get} API initiates a multi-stage resolution process. The receiver queries the resident daemon to obtain the physical location and device ID of heterogeneous objects. 
As a portability layer, the \texttt{Set}/\texttt{Get} API decouples logical data operations from physical transport backends.
The available transport backend is selected based on the physical location and hardware stacks.
Once the peer identity is established, the sender/receiver instantiates or reuses a specialized communication domain to execute point-to-point primitives (e.g., HCCS/NVLink backend).
Completing communication over different accelerators may change specific transfer latency, instead of graph-driven scheduling semantics.

H2D/D2H transfer is primarily used by the Training Engine to swap weights and optimizer states for suspended and resumed agents. The \texttt{Set} API instantiates a buffer in the host memory. The resident daemon then registers metadata that links the device memory extents to the corresponding host-side buffer. Following metadata registration, heterogeneous objects are asynchronously migrated from the device to the host memory via D2H transfer.
When a receiver initiates a \texttt{Get} request, the location metadata can be queried from the resident daemon using the user-defined key. The heterogeneous objects are directly copied from the local host memory to the target device. 
In the case of cross-node retrieval, the resident daemon synchronizes the remote sender and the local receiver to facilitate a high-throughput, zero-copy migration via an RDMA-enabled network \cite{lu2025fast}.
Once the data is staged within the local host domain, it is finalized through a remote Host-to-Device (RH2D) operation.

\subsection{Practical Deployment Lessons}
Building FlexMARL and deploying it on a production cluster provided unique insights into the LLM-based MARL training. We summarize our experiences into the following key lessons during MARL system development.

\textbf{Hardware-Aware Abstraction.} 
Modern distributed frameworks usually abstract away physical device details to simplify development. 
However, the absence of stable UUIDs in the NPU driver stack led to failures in high-level zero-copy mechanisms. FlexMARL addresses this by enforcing process-level determinism and mapping internal PIDs to logical ranks. Such an explicit addressing scheme prevents heterogeneous objects from depending on unstable driver-level identifiers. 
It is worth noting that dependency parallel sampling, hierarchical load balancing, agent-level asynchronous pipeline, and on-demand resource allocation are hardware-agnostic. The core designs in FlexMARL can apply to other accelerators. The NPU-specific implementation aims to reflect practical deployment lessons, not fundamental design requirements.

\textbf{Rollout-Training Parameter Synchronization.} 
We find that existing fine-grained, parameter-by-parameter synchronization schemes are costly. The task scheduling and kernel launching overhead accounts for over 99\% of synchronization latency when iterating over billions of parameters. 
This suggests that the cost of invoking communication primitives could dwarf the actual time spent on transferring data.
FlexMARL aggregates weights into a single contiguous memory buffer, reducing synchronization complexity from $O(Num\_of\_Parameters)$ to $O(1)$ and achieving a $200\times$ speedup in practical deployment. 
These experiences reveal that system software for accelerators must aggressively batch control-plane operations to amortize scheduling costs.

\textbf{Cross-Node Agent Deployment.} 
Agent deployment in most existing frameworks fails to scale effectively across multiple nodes, revealing critical performance bottlenecks. 
The reason is that all resources across nodes are managed via a single placement group (PG) using a ``PACK'' scheduling strategy\footnote{https://docs.ray.io/en/latest/ray-core/scheduling/placement-group.html}. 
Both training and inference processes are deployed in the same PG. 
By generating a bundle list derived from a predefined device set, each process is bound to its corresponding device bundle.
However, since the logical bundle ordering is not equivalent to physical device IDs, the ``PACK'' strategy fails in multi-agent settings. Training/inference processes of the same agent may be scheduled to different nodes, leading to additional communication overhead and potential network failure risk \cite{moritz2018ray}.
By shifting scheduling granularity from a cluster level to individual nodes, our framework instantiates an independent PG for each node using a ``STRICT\_PACK'' strategy\footnotemark[1]. 
By establishing a deterministic one-to-one mapping between logical bundles and physical device IDs, FlexMARL effectively eliminates unintended cross-node communication and enhances system stability during large-scale MARL deployment.

%% file: content/evaluation.tex
Our evaluation aims to answer the following six research questions (RQ):
\begin{itemize}
\setlength{\itemsep}{1pt}
\setlength{\parsep}{1pt}
\setlength{\parskip}{1pt}
    \item \textbf{RQ1}: How does FlexMARL compare with existing MARL frameworks in end-to-end training performance?
    \item \textbf{RQ2}: Can FlexMARL mitigate topology-induced load imbalance in multi-agent interactions?
    \item \textbf{RQ3}: Can FlexMARL improve hardware utilization for dynamic resource demands?
    \item \textbf{RQ4}: Are the key components of FlexMARL necessary for improving MARL training efficiency?
    \item \textbf{RQ5}: What is the control plane overhead of FlexMARL?
    \item \textbf{RQ6}: How does FlexMARL perform with increasing agent counts and model sizes?
\end{itemize}

\begin{table*}[t]
\centering
\caption{Overall training performance of different MARL frameworks on four datasets.}
\label{tab:overall_performance}
\setlength{\tabcolsep}{2.5pt}
\renewcommand{\arraystretch}{1.1}
\resizebox{\textwidth}{!}{
\begin{tabular}{l|ccc|ccc|ccc|ccc}
\toprule
\multirow{2}{*}{\centering\textbf{Framework}}
& \multicolumn{3}{c|}{\textbf{GSM8K}}
& \multicolumn{3}{c|}{\textbf{ToolBench}}
& \multicolumn{3}{c|}{\textbf{MA}}
& \multicolumn{3}{c}{\textbf{CA}} \\ \cline{2-13}
& E2E Time & Speedup & Throughput
& E2E Time & Speedup & Throughput
& E2E Time & Speedup & Throughput
& E2E Time & Speedup & Throughput \\
\midrule
MAS-RL
& 119.3s & 1.0$\times$ & 66.4tps
& 404.3s & 1.0$\times$ & 189.8tps
& 914.4s & 1.0$\times$ & 119.0tps
& 438.6s & 1.0$\times$ & 265.5tps \\

DistRL
& 39.6s & 3.0$\times$ & 298.0tps
& 98.5s & 4.1$\times$ & 945.1tps
& 293.8s & 3.1$\times$ & 401.0tps
& 130.0s & 3.4$\times$ & 571.6tps \\

MARTI
& 35.6s & 3.4$\times$ & 314.0tps
& 87.9s & 4.6$\times$ & 1123.0tps
& 174.1s & 5.3$\times$ & 642.8tps
& 112.8s & 3.9$\times$ & 655.9tps \\

\rowcolor{gray!20}
\textbf{FlexMARL}
& \textbf{26.3s} & \textbf{4.5$\times$} & \textbf{364.9tps}
& \textbf{60.0s} & \textbf{6.7$\times$} & \textbf{1579.4tps}
& \textbf{126.1s} & \textbf{7.3$\times$} & \textbf{910.2tps}
& \textbf{78.8s} & \textbf{5.6$\times$} & \textbf{821.4tps} \\
\bottomrule
\end{tabular}}
\vspace{1mm}
\end{table*}

\subsection{Experimental Setup}

\textbf{Models and Datasets.}
We validate the effectiveness of FlexMARL on two public datasets and two industrial datasets.
The public datasets include two representative agentic tasks.
GSM8K \cite{cobbe2021training} focuses on mathematical reasoning, and ToolBench \cite{qin2024toolllm} evaluates tool-use capabilities.
However, these open-source benchmarks typically involve fewer agents, smaller models, and simpler interaction patterns, making them insufficient to capture the infrastructure bottlenecks in large-scale, real-world MARL deployments.
Therefore, we additionally introduce two real-world industrial datasets, Merchant Assistant (MA) and Category Assistant (CA).
For the MA dataset, multiple agents collaborate to help merchants perform complex store management tasks, including sales performance analysis, marketing strategy optimization, and after-sales service. 
For the CA dataset, MAS is responsible for order querying, pricing strategies, and inventory management suggestions. 
The above production datasets are constructed for specific e-commerce tasks to simulate multi-agent collaboration requirements. Each sample comprises structured user queries, historical interaction context, and ground-truth agent responses.
The detailed information about the MA and CA datasets is hidden due to business and confidentiality concerns.
We employ the Qwen2.5-14B model as the backbone for the GSM8K, ToolBench, and MA datasets, in which each agent does not share parameters and independently optimizes its policy. The policy models for the CA dataset involve Qwen2.5-14B and Qwen2.5-32B.

\textbf{Training Configurations.}
For dependency parallel sampling, inter-query and intra-query parallelism are set to 4 and 16, respectively, to accelerate the processing of user queries. The maximum length of response tokens is 8192. 
The load-balancing disparity threshold $\Delta$ is set to 5. The batch size and micro batch size in our agent-level asynchronous pipeline are set to 64 and 16, respectively. 
We adopt the GRPO algorithm \cite{shao2024deepseekmath} for policy training and use a learning rate of 1e-6 with the Adam optimizer.
All experiments are run with a fixed random seed of 2048 for reproducibility.

\textbf{Baselines.}
Most existing RL frameworks \cite{hu2024openrlhf,sheng2025hybridflow,fu2025areal} are designed for a single model and cannot support MARL. 
Under such circumstances, we compare our proposed framework against the following MARL baselines:
\begin{itemize}
    \item \textbf{MAS-RL} extends the single-agent RL training framework to multi-agent settings, reflecting standard practice without specialized MARL infrastructure. The naive MARL implementation adopts the \textit{colocated} architecture and performs uniform scheduling and static resource allocation during both rollout and training phases.
    \item \textbf{DistRL} enables flexible resource allocation through the \textit{disaggregated} architecture, avoiding frequent onload/offload operations. The rollout and training phases of MARL are deployed on dedicated resource pools and coordinated under a synchronous paradigm.
    \item \textbf{MARTI} \cite{marti2025} is the state-of-the-art MARL framework that supports centralized multi-agent interactions and distributed policy training. Asynchronous rollouts are designed to further improve MARL training efficiency, following its official implementation.
\end{itemize}
For fair comparison, the baselines are implemented and tuned under the same total number of resources, model implementation, GRPO training configuration, global/micro batch sizes, and optimizer parameters.

\textbf{Evaluation Metrics.} The design goal of FlexMARL is to facilitate efficient training of MARL algorithms on large-scale infrastructure. 
Our experiments focus on training efficiency metrics to comprehensively evaluate performance. (1) E2E time records the average latency of all phases (rollout, training, and others) for a single sample. (2) Speedup is calculated as the ratio of the E2E time between frameworks and MAS-RL, quantifying the relative performance improvement. (3) Throughput is denoted by the number of generated tokens per second. (4) Task accuracy is used to verify whether our solution preserves the benefit of MARL policy optimization.
(5) Agent load is reflected by the total number of rollout requests that each agent has processed. (6) Utilization rate represents the percentage of time that AI cores remain active within a given training period.
This AI Core active time metric on NPUs measures actual matrix computations and covers only a subset of resource utilization metrics.
Considering different accounting rules across various hardware stacks, we focus on relative utilization improvement rather than absolute values.

\begin{figure}[t]
\centering
\subfigure[MA Dataset]{
\centering
\includegraphics[width=0.472\linewidth]{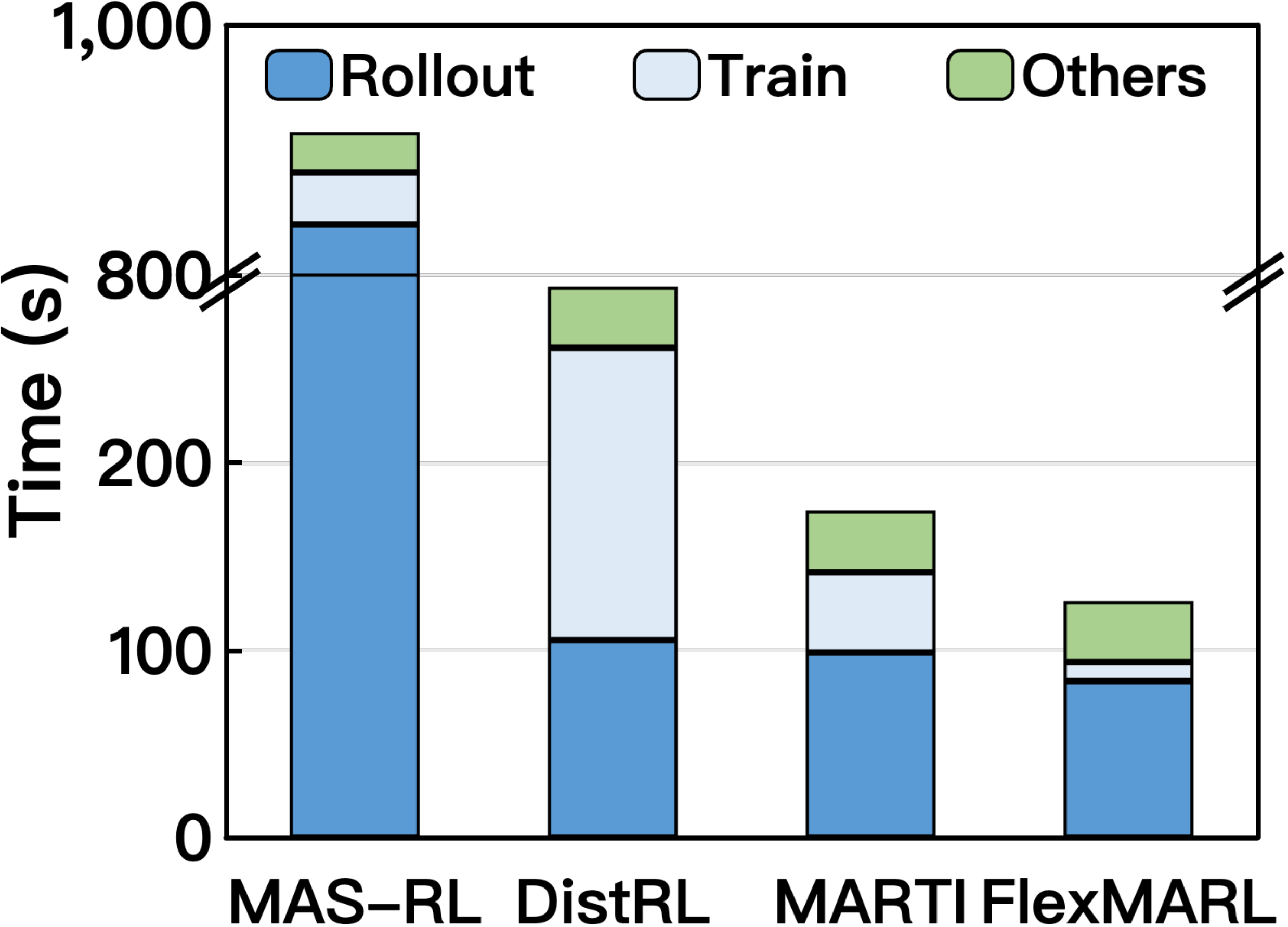}
}
\subfigure[CA Dataset]{
\centering
\includegraphics[width=0.472\linewidth]{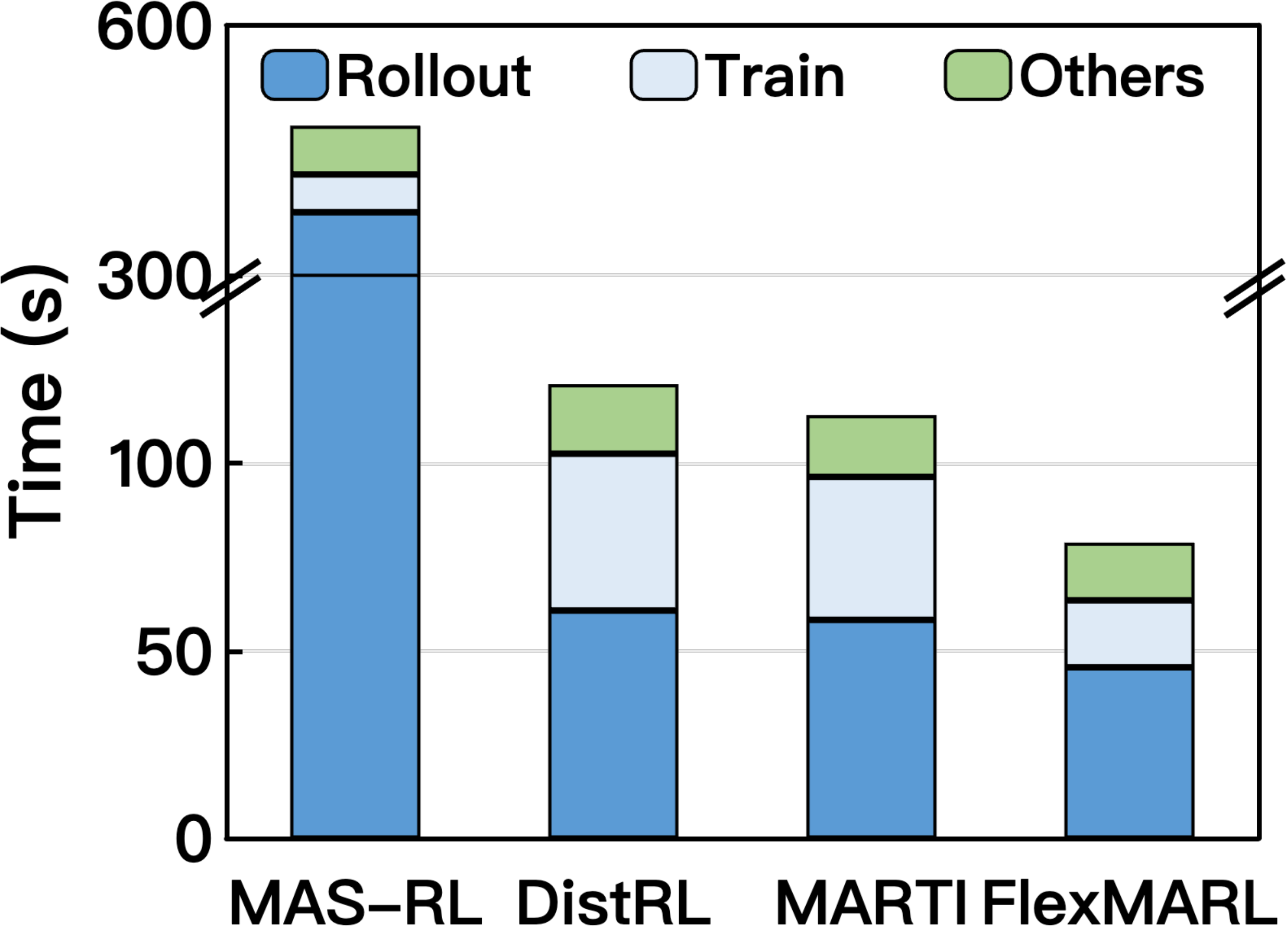}
}
\centering
\vspace{-2mm}
\caption{E2E time breakdown of different MARL frameworks on the MA and CA datasets.}
\label{fig:time_breakdown}
\end{figure}

\subsection{RQ1: Overall Performance}
We first evaluate the overall performance of FlexMARL and three representative baselines. Table \ref{tab:overall_performance} summarizes the overall training efficiency on four datasets. Figure \ref{fig:time_breakdown} shows the performance breakdown during rollout and training phases on the MA and CA datasets. FlexMARL consistently delivers the highest training speed and throughput. 
We highlight three observations from the experimental results. 
Firstly, our proposed framework outperforms MAS-RL by a large margin. For example, FlexMARL simultaneously reduces rollout and training time, ultimately achieving end-to-end speedups of 4.5$\times$, 6.7$\times$, 7.3$\times$, and 5.6$\times$ on the GSM8K, ToolBench, MA, and CA datasets. The corresponding throughput improvements are 5.5$\times$, 8.3$\times$, 7.6$\times$, and 3.1$\times$, respectively. The naive adaptation in MAS-RL cannot address MARL-specific challenges and suffers from severe queuing delays. FlexMARL leverages parallel sampling and elastic scaling to eliminate such bottlenecks.
Secondly, even with the disaggregated architecture, the E2E speed of DistRL is only 42\%–67\% of ours. Compared to DistRL, we observe from Figure \ref{fig:time_breakdown} that FlexMARL reduces policy optimization time from 155.9s to 10.2s on the MA dataset.
DistRL relieves resource contention but introduces significant pipeline bubbles, resulting in resource waste.
Our agent-level asynchronous pipeline overlaps rollout and training phases, hiding tail latency. 
Thirdly, for public and industrial datasets, FlexMARL improves end-to-end speed and throughput by 1.4$\times$ and 1.3$\times$ over the state-of-the-art MARTI framework on average. Although asynchronous rollouts can speed up experience collection, their fixed resource allocation strategy fails to accommodate skewed and dynamic workloads. 
The results above demonstrate that our co-design of the rollout, training, and orchestration phases provides substantial performance gains.

\begin{figure}[t]
\centering
\subfigure[Agent A]{
\centering
\includegraphics[width=0.46\linewidth]{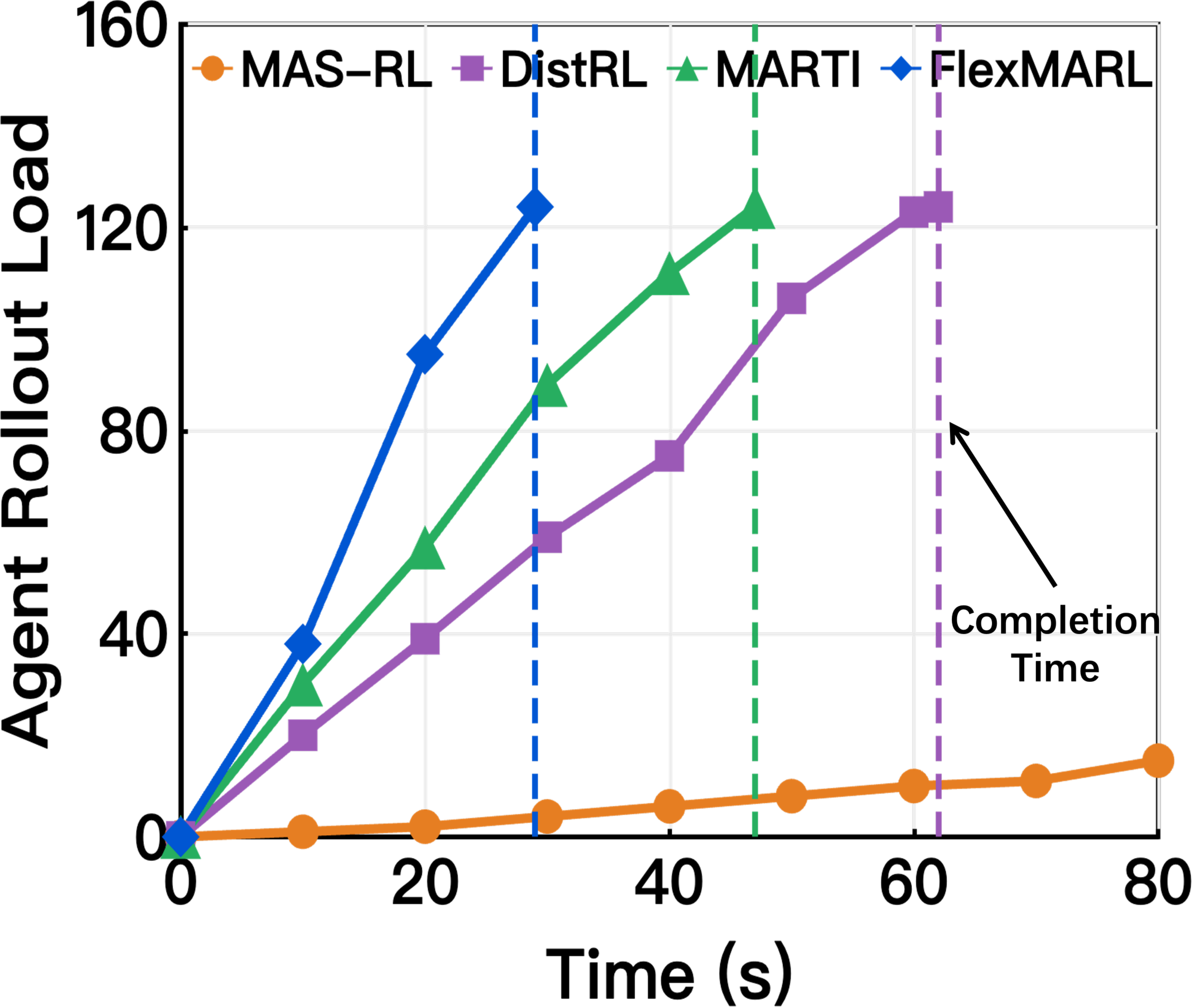}
}
\subfigure[Agent B]{
\centering
\includegraphics[width=0.47\linewidth]{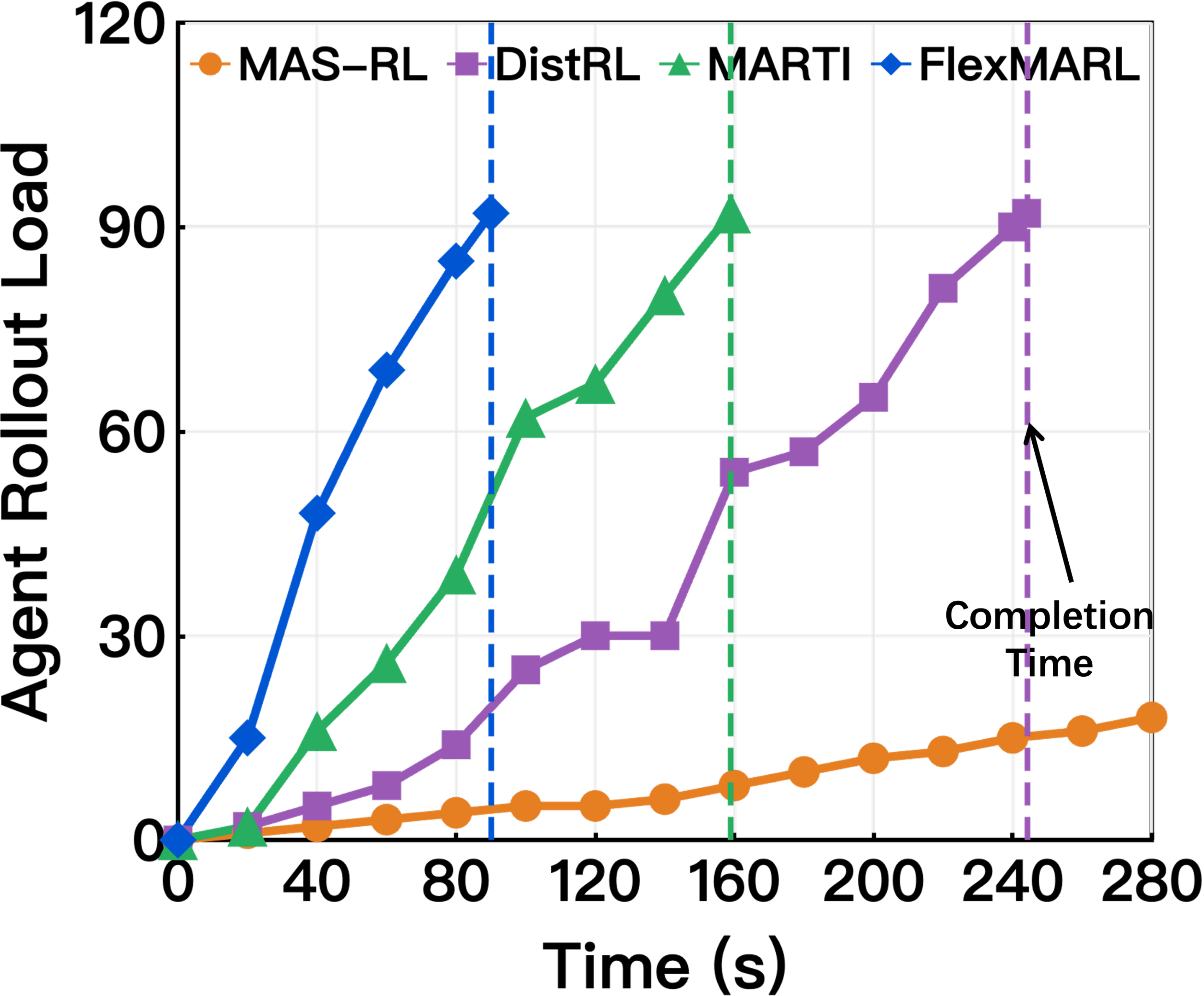}
}
\vspace{-3mm}
\caption{Processed rollout load of representative agents on the MA dataset.}
\label{fig:ma load}
\vspace{-2mm}
\end{figure}

\begin{figure}[t]
\centering
\subfigure[Agent A]{
\centering
\includegraphics[width=0.47\linewidth]{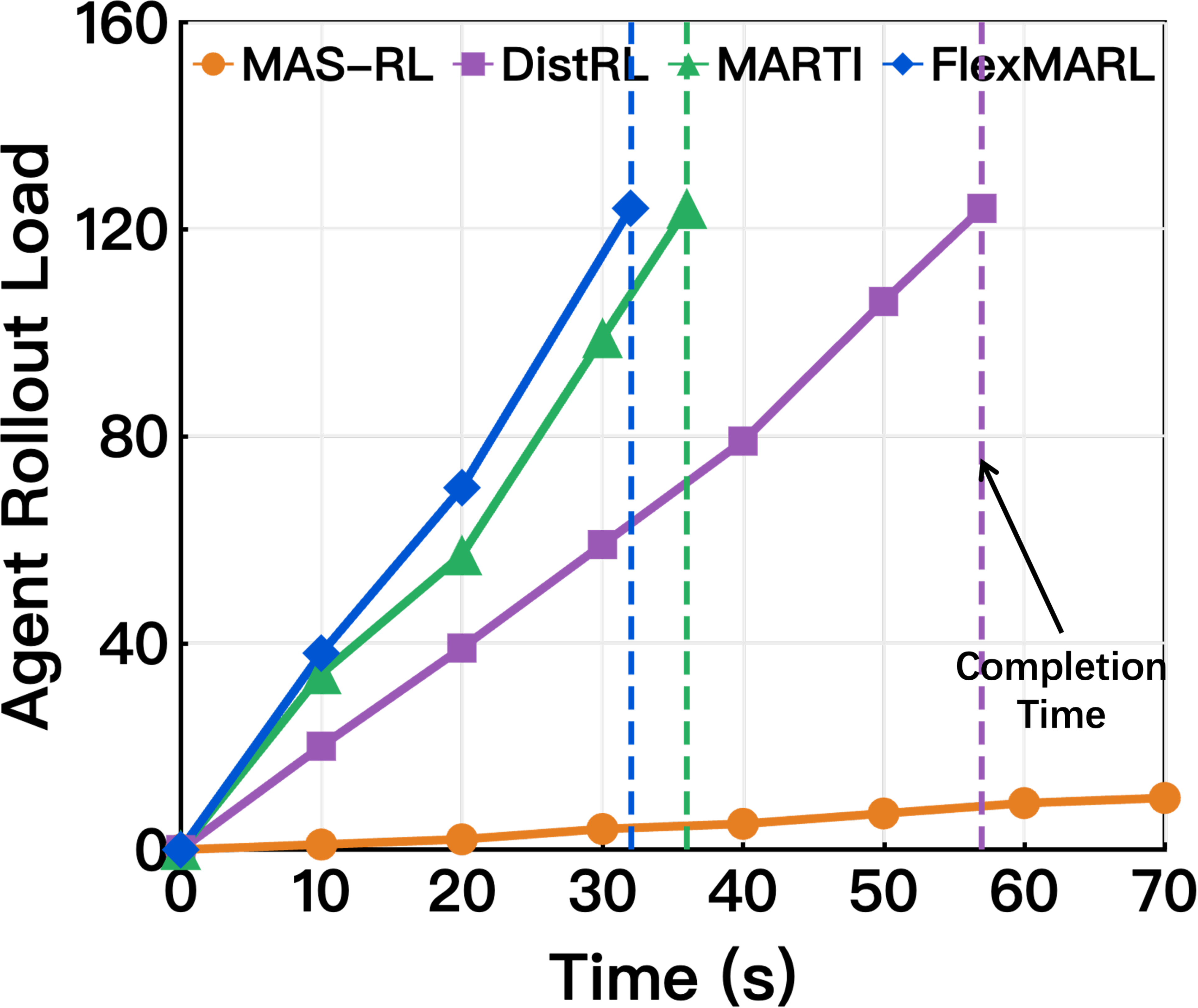}
}
\subfigure[Agent B]{
\centering
\includegraphics[width=0.47\linewidth]{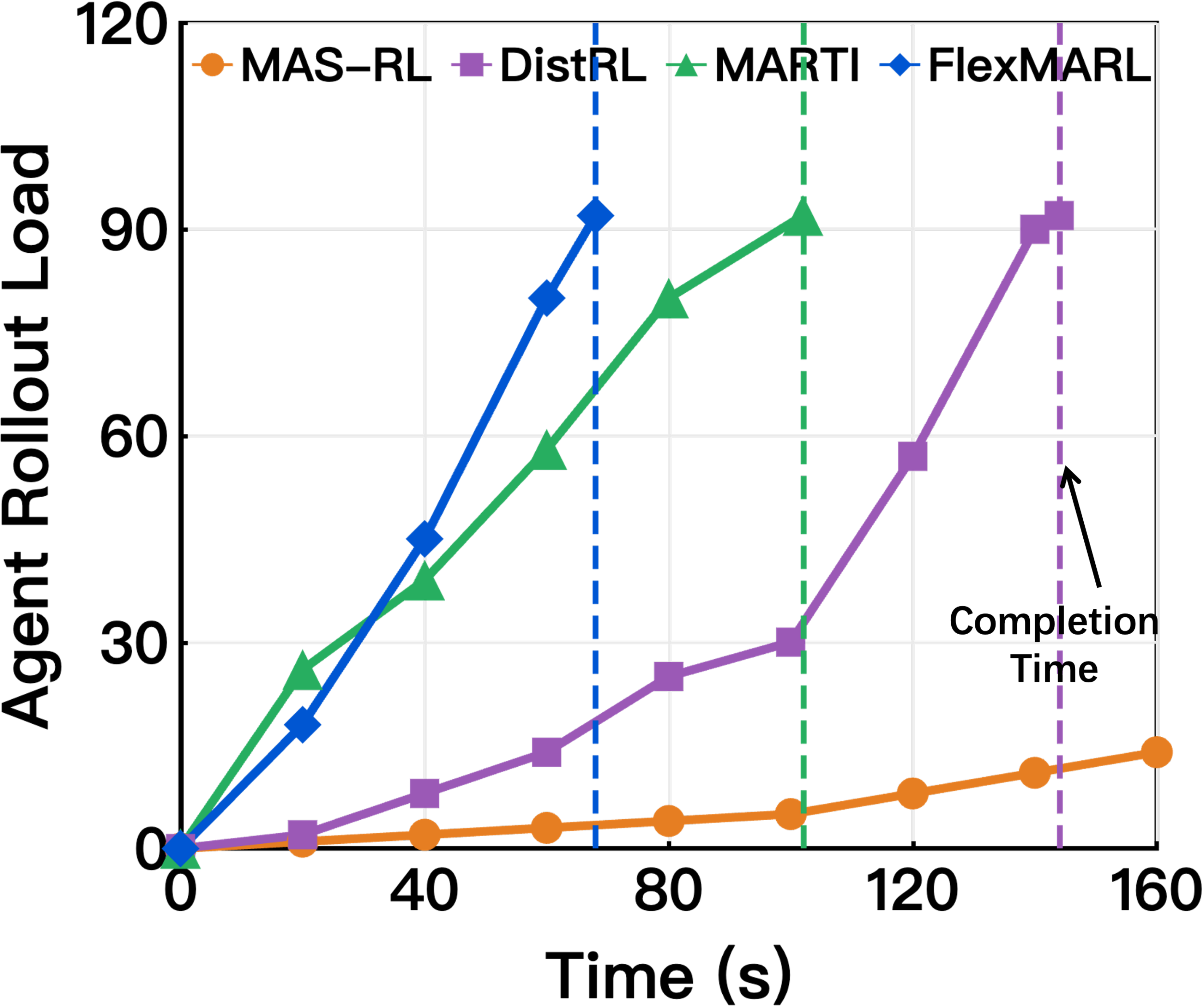}
}
\vspace{-3mm}
\caption{Processed rollout load of representative agents on the CA dataset.}
\label{fig:ca load}
\vspace{-2mm}
\end{figure}

\subsection{RQ2: Agent Rollout Load}
To validate the rollout efficiency, we select two representative agents from two datasets and monitor the number of processed requests within a single step. Figures \ref{fig:ma load} and \ref{fig:ca load} plot the load patterns of different frameworks. We find that FlexMARL processes rollout requests much faster than the baselines for all datasets. For instance, FlexMARL completes all pending requests within 90s for Agent B on the MA dataset, while DistRL and MARTI require about 244s and 159s. 
The corresponding speedups are 2.7$\times$ and 1.8$\times$, respectively. In particular, MAS-RL maintains a constant workload throughout training, as it processes all queries serially without parallelization, resulting in the longest completion time. These baselines cannot mitigate topology-induced rollout skew.
The core agents experience persistent queueing of rollout requests, which blocks the MARL training process. 
Compared to other methods, the queued requests of FlexMARL exhibit a faster increase followed by quicker decrease. 
The explanation for this phenomenon is that our Rollout Engine combines dependency parallel sampling with elastic scaling, achieving balanced workloads across multiple agents. 
Upstream agents also adjust the number of inference instances to efficiently process their queued requests, thereby accelerating rollout along the MEG.

\begin{figure}[t]
\centering
\subfigure[MA Dataset]{
\centering
\includegraphics[width=0.47\linewidth]{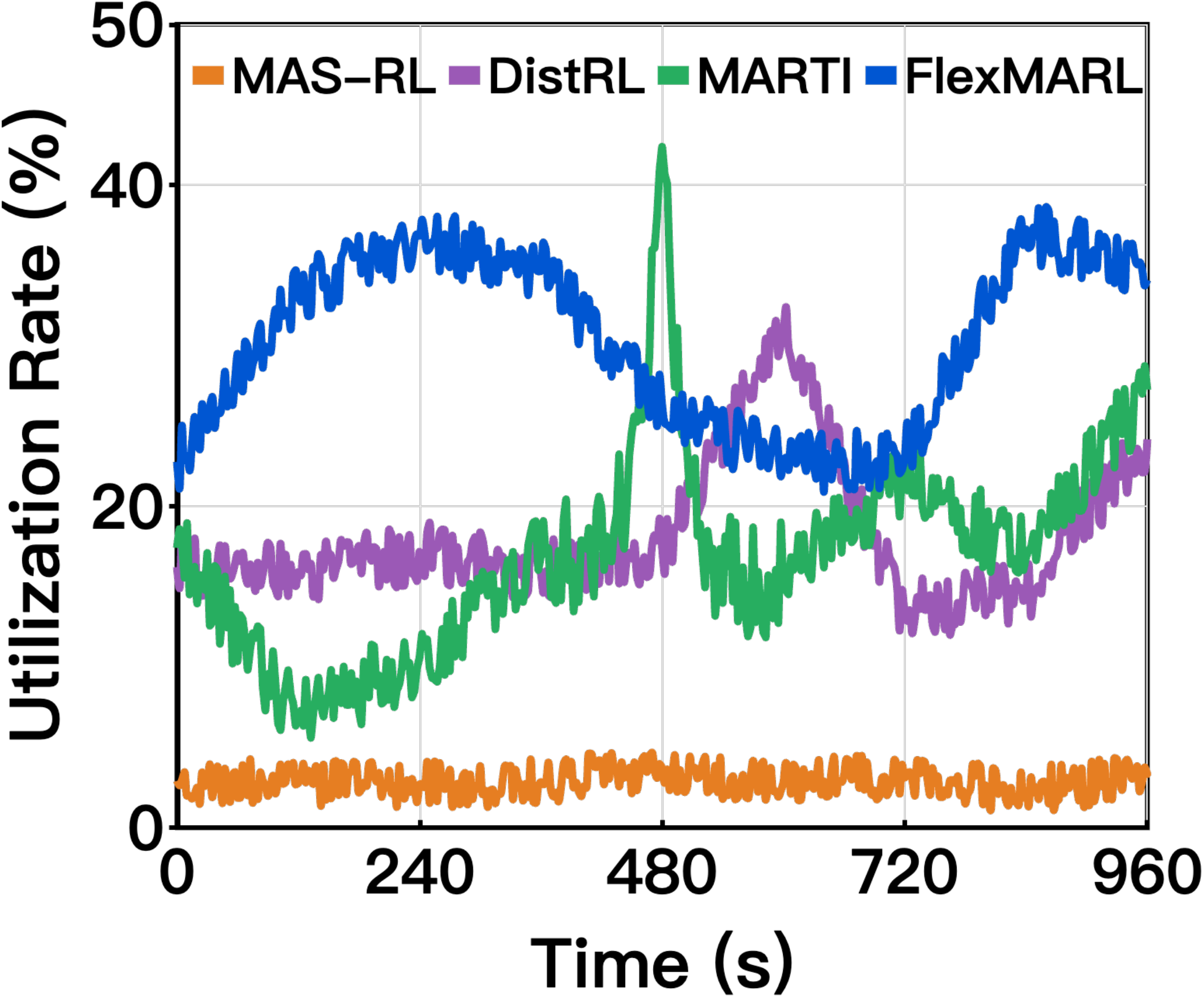}
}
\subfigure[CA Dataset]{
\centering
\includegraphics[width=0.47\linewidth]{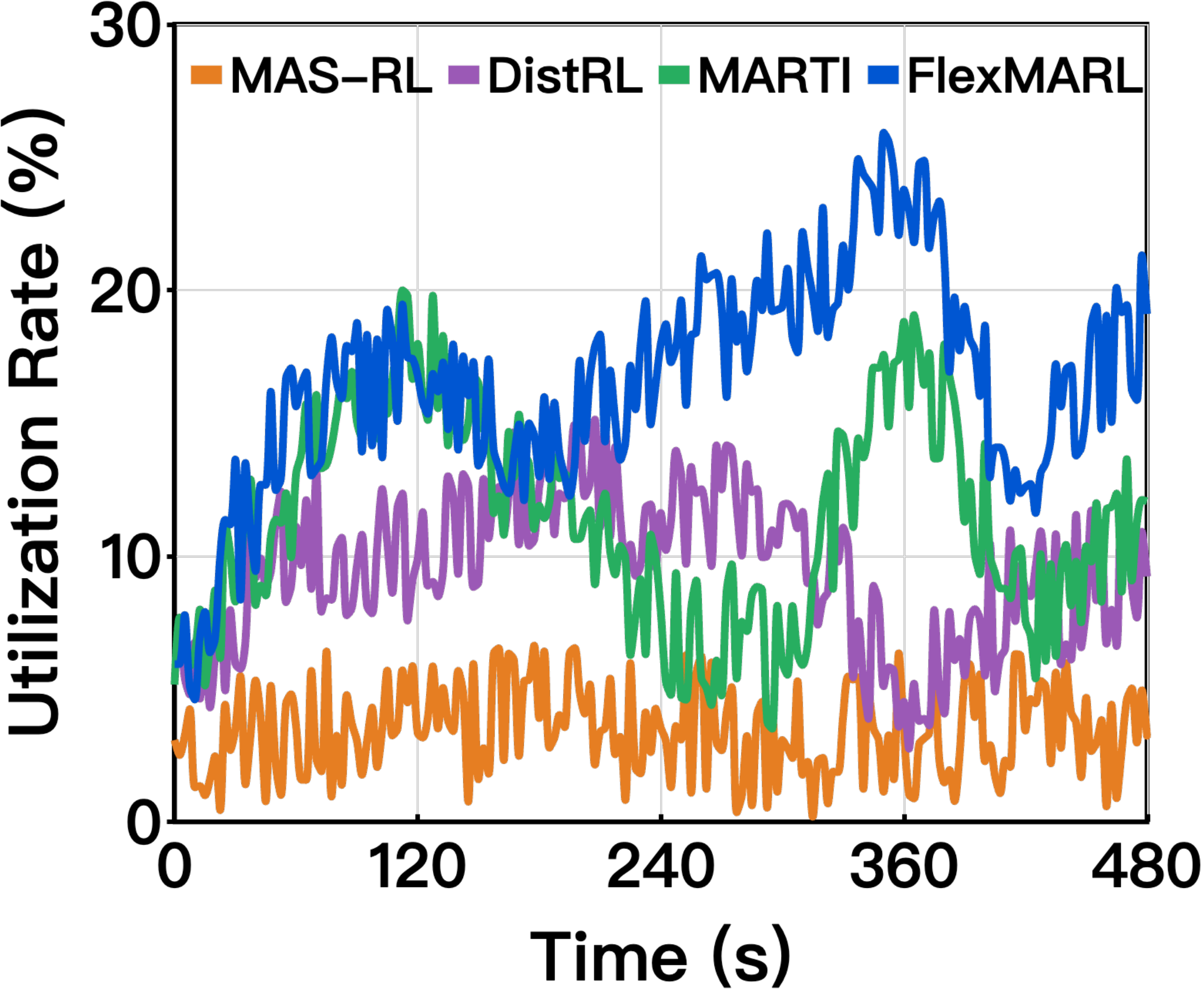}
}
\vspace{-2mm}
\caption{Resource utilization rates of different MARL frameworks on the MA and CA datasets.}
\label{fig:utilization rates}
\end{figure}

\subsection{RQ3: Resource Utilization}
We measure the overall utilization rates of FlexMARL and three baselines throughout the MARL training process. 
Resource allocations are already matched for different systems. 
The results on the MA and CA datasets are presented in Figure \ref{fig:utilization rates}. Compared with both colocated and disaggregated baselines, FlexMARL achieves a substantial relative improvement in resource utilization rates. Other frameworks struggle with inefficient resource usage. For example, MAS-RL and MARTI exhibit severe utilization gaps with an average of only 3.6\% and 12.3\% on the CA dataset. FlexMARL provides 5.6$\times$ and 1.6$\times$ utilization improvement, respectively. Their colocated architecture forces training and rollout to use the same number of resources. 
Heterogeneous requirements of different phases result in inherent resource conflict. 
Furthermore, compared with already-disaggregated DistRL, resource utilization rates achieved by FlexMARL still increase by nearly 2$\times$ on the MA and CA datasets.
This is because the rollout and training phases are executed alternately in DistRL, which leaves training resources idle and shows noticeable utilization drops. 
The above results imply that disaggregation itself is beneficial, while agent-level asynchronous pipeline and on-demand resource allocation can accommodate dynamic training demand, further improving resource utilization.

\begin{table}[t]
\centering
\caption{Ablation study of core components in FlexMARL.}
\label{tab:ablation_study2}
\resizebox{0.5\textwidth}{!}{
\begin{tabular}{c |l |ccc}
\toprule
Dataset & Variant & E2E Time & Speedup & Throughput\\
\midrule
\multirow{4}{*}{MA} & w/o balancing      &152.2s &6.0$\times$ & 729.9tps \\
                    & w/o async      & 256.2s & 3.6$\times$ & 444.0tps  \\
                    & w/o allocation & 132.2s & 6.9$\times$ & 784.7tps \\

                    &\cellcolor{gray!20}FlexMARL &\cellcolor{gray!20}126.1s &\cellcolor{gray!20}7.3$\times$ &\cellcolor{gray!20}910.2tps \\
\midrule
\multirow{4}{*}{CA} & w/o balancing     &95.9s &4.6$\times$ & 611.8tps \\
                    & w/o async       & 124.1s & 3.5$\times$ & 608.4tps \\
                    & w/o allocation & 83.6s & 5.2$\times$ & 759.1tps \\

                    &\cellcolor{gray!20}FlexMARL &\cellcolor{gray!20}78.8s &\cellcolor{gray!20}5.6$\times$ &\cellcolor{gray!20}821.4tps  \\
\bottomrule
\end{tabular}
}
\end{table}

\begin{table}[t]
\centering
\caption{Task accuracy comparison under synchronous and asynchronous pipelines.}
\label{tab:acc}
\resizebox{0.38\textwidth}{!}{
\begin{tabular}{c | ccc}
\toprule
Dataset & SFT-only & Sync MARL & FlexMARL \\
\midrule
MA & 86.3\% &90.3\% & 90.1\%  \\
CA & 86.4\% &90.4\% & 90.6\% \\
\bottomrule
\end{tabular}}
\vspace{1mm}
\end{table}

\subsection{RQ4: Ablation Study}
We conduct ablation studies to quantify the individual contributions of rollout load balancing, agent-level asynchronous pipeline, and on-demand resource allocation. 
As summarized in Table \ref{tab:ablation_study2}, removing load balancing reduces the training throughput by 19.8\%-25.5\%, and the speedup drops to 6.0$\times$ and 4.6$\times$. 
This indicates that our Rollout Engine elastically reallocates inference capacity and prevents overloaded agents from becoming persistent bottlenecks. 
The w/o async variant shows how a well-tuned synchronous system would actually perform against FlexMARL. 
It retains load balancing and on-demand allocation, while only disabling the asynchronous pipeline.
Even against this strong synchronous design, FlexMARL still reduces E2E time from 256.2s to 126.1s (2.0$\times$) on MA and from 124.1s to 78.8s (1.6$\times$) on CA, highlighting the necessity of our agent-level asynchronous pipeline.
The Joint Orchestrator allows training to begin as soon as part of the experience is available and effectively hides the long-tail latency of critical paths.
Agent-centric resource allocation in the Training Engine is also important for adapting to dynamic agent workloads. Without the allocation component, idle resources cause a decrease in training speed.
These results indicate that the above core components are necessary for achieving the full performance of FlexMARL.

We further verify that our training efficiency gains do not come at the cost of consistency.
Table \ref{tab:acc} reports the final task accuracy of FlexMARL, synchronous MARL baseline, and supervised fine-tuning (SFT) baseline.
We observe that FlexMARL maintains task accuracy nearly identical to the synchronous baseline.
Meanwhile, both synchronous MARL and FlexMARL achieve approximately 4\% accuracy improvement compared with the SFT-only baseline, demonstrating the benefit of MARL policy optimization.

\begin{figure}[t]
\centering
\subfigure[Swap-in Overhead]{
\centering
\includegraphics[width=0.47\linewidth]{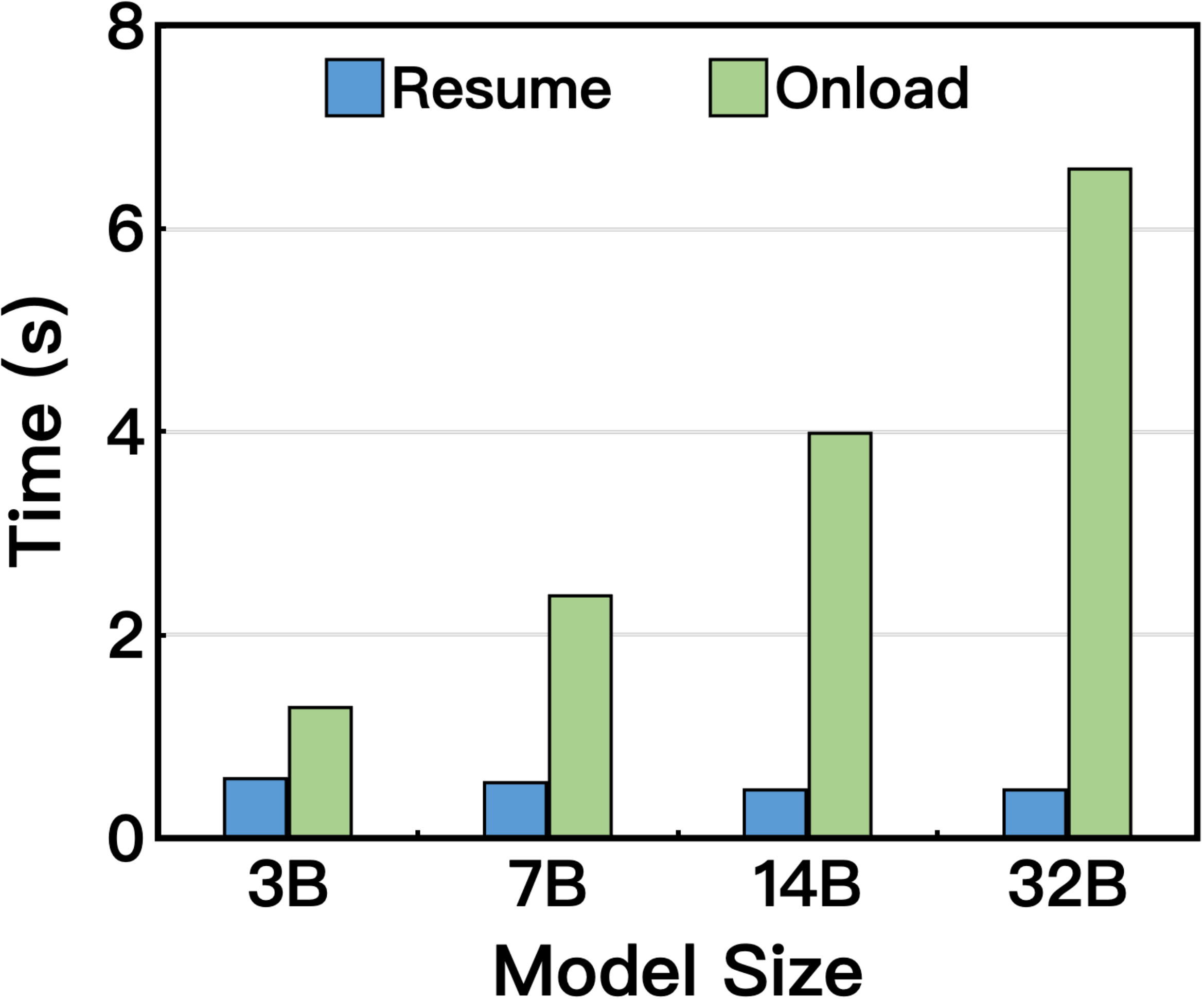}
}
\subfigure[Swap-out Overhead]{
\centering
\includegraphics[width=0.47\linewidth]{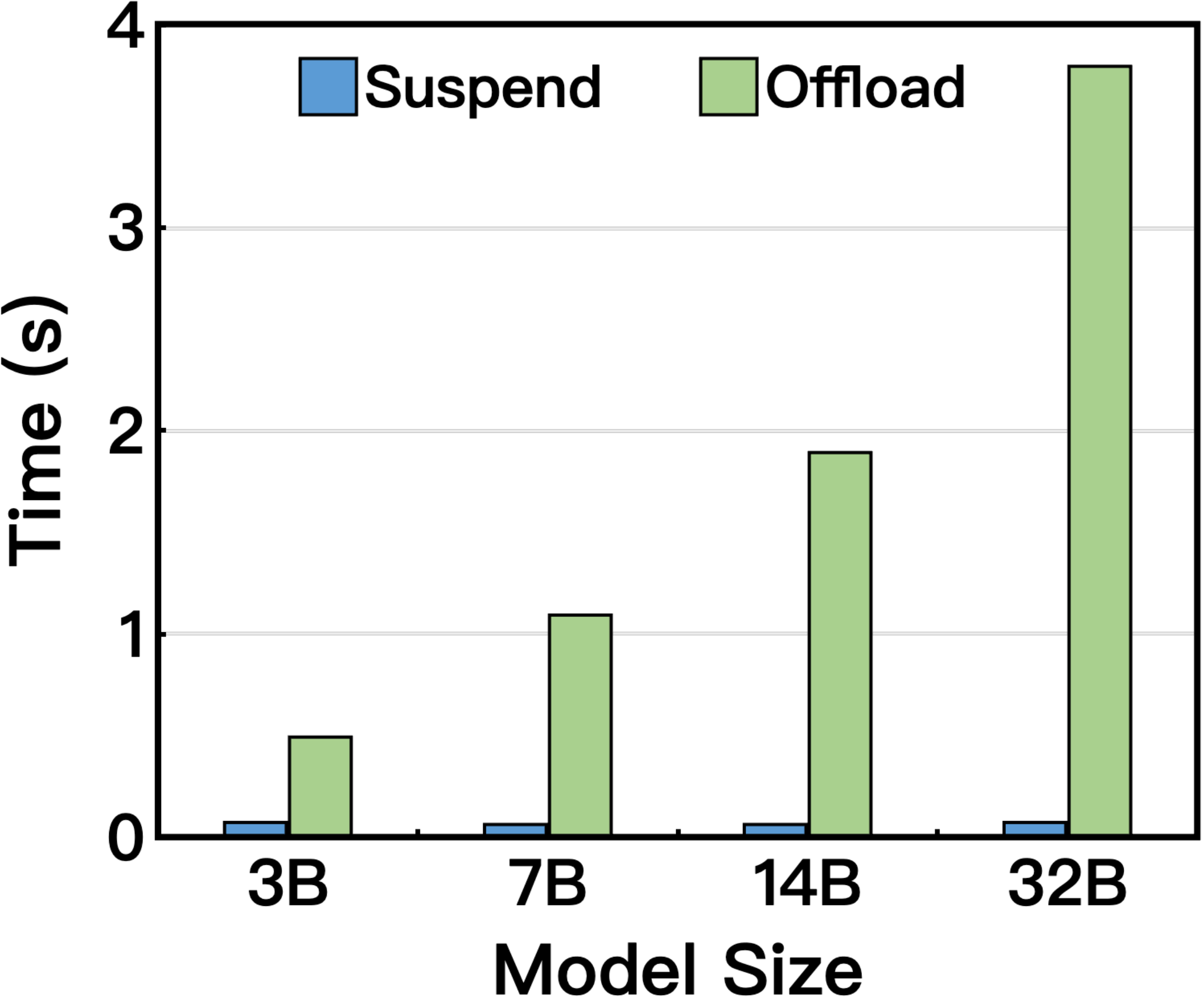}
}
\vspace{-2mm}
\caption{Swap-in/out overhead of training states across different model sizes.}
\label{fig:swap overhead}
\end{figure}

\subsection{RQ5: Control Plane Overhead}
As a distributed system, FlexMARL incurs overhead from state communication and control plane operations. We conduct high-load tests to verify that they do not become system bottlenecks. We first quantify the state swap overhead across four representative model sizes.
The swap-in overhead includes resuming the process group and loading states back into device memory via \texttt{Get} API.
The swap-out overhead includes suspending the process group and offloading training states to host memory via \texttt{Set} API.
As shown in Figure \ref{fig:swap overhead}, both suspend and resume latencies are minimal and stay nearly constant regardless of model scale. The onload and offload overhead grow gradually with model size. 
Nevertheless, after integrating the location-agnostic \texttt{Set}/\texttt{Get} API, total swap overhead is only 11s for the largest model, which can be effectively hidden behind rollout latency.

We also measure the latency of other core operations under production workload conditions in Table \ref{tab:overhead}. Load monitoring in the Rollout Engine involves periodic polling of request queue lengths across all agents, incurring a latency of 380$\sim$610$\mu$s. For balancing decision-making, which performs load disparity calculation and elastic scaling decisions, the latency is only 50$\sim$100$\mu$s even for 1,000 agents. 
When the number of agents increases to 8,192 agents, concurrent experience retrieval requests in the MEG store take 6.7ms within one node. For inter-node service, 8,192 concurrent requests achieve 25.0ms average latency, demonstrating scalability beyond the evaluated number of training agents.
These results verify that additional overheads introduced by FlexMARL are acceptable compared to the E2E training time, enabling large-scale deployments of LLM-based MARL.

\begin{table}[t]
\centering
\caption{Control plane overhead breakdown of FlexMARL.}
\label{tab:overhead}
\vspace{-1mm}
\resizebox{0.48\textwidth}{!}{
\begin{tabular}{l |c}
\toprule
Operation & Control Plane Overhead \\
\midrule
Load Monitoring & 380$\sim$610$\mu$s \\
Scaling Decision-making & 50$\sim$100$\mu$s \\
Intra-node MEG Store Service & 6.7ms  \\
Inter-node MEG Store Service & 25.0ms  \\
\bottomrule
\end{tabular}}
\vspace{1mm}
\end{table}

\subsection{RQ6: Scalability}

We proceed to investigate the system performance of FlexMARL under large-scale heterogeneous scenarios. 
Table \ref{tab:large_scale_results} reports the E2E time and training throughput on the MA dataset across varying agent counts and model sizes. FlexMARL demonstrates robust scalability, achieving a peak throughput of 754.2tps with 15 agents. Critically, FlexMARL successfully supports complex heterogeneous configurations, such as a mixed multi-agent setting of 3$\times$32B and 7$\times$14B. 
Such configurations pose significant challenges to most existing MARL frameworks \cite{marti2025}, which lack cross-node agent placement capabilities required for such heavy, uneven workloads. Consequently, these frameworks cannot accommodate individual agents whose resource requirements exceed the capacity of a single node, and cannot be directly compared at heterogeneous 14B/32B scales. In contrast, our end-to-end optimization enables FlexMARL to maintain high throughput even for large model sizes and production cluster scales, not merely agent count.
The above results demonstrate our ability to efficiently orchestrate heterogeneous multi-agent workloads at a scale beyond that practically evaluated in prior LLM-based MARL systems.

%% file: content/related_work.tex
\textbf{Frameworks for LLM Post-Training.}
RL has emerged as a critical paradigm for enhancing LLM capabilities \cite{guo2025deepseek,yu2025dapo}, and it relies heavily on scalable training infrastructure \cite{zhang2025disttrain,meng2025astral,ge2025bytescale}. Early frameworks like ColossalChat \cite{li2023colossal} and DeepSpeed-Chat \cite{yao2023deepspeed} adopt the colocated architecture where the rollout and training phases share the same resource pool, leading to resource contention. 
To improve throughput, recent RL frameworks such as OpenRLHF \cite{hu2024openrlhf}, veRL \cite{sheng2025hybridflow}, and AReaL \cite{fu2025areal} enable asynchronous pipeline and leverage distributed runtimes \cite{moritz2018ray,chen2024yuanrong} to orchestrate the different phases.
AgentRL \cite{zhang2025agentrl} and SkyRL-Agent \cite{cao2025skyrl} further extend these pipelines to support multi-turn and long-horizon agentic RL training.
Other solutions aim to improve inference efficiency by using quantized actors \cite{li2026qurl} or adaptive rollout allocation \cite{zhang2026train}.
However, these general-purpose RL frameworks are primarily designed for a single policy model and cannot capture cross-agent dependencies, agent load skew, or dynamically activated training subsets inherent in LLM-based MARL.

\begin{table}[t]
\centering
\caption{Performance evaluation of FlexMARL in large-scale heterogeneous MARL scenarios.}
\label{tab:large_scale_results}
\vspace{-1mm}
\setlength{\tabcolsep}{0pt}
\begin{tabular*}{\columnwidth}{@{\extracolsep{\fill}} l| cccc}
\toprule
Configuration & Rollout & Training & E2E Time & Throughput \\
\midrule
5$\times$32B & 96.2s &54.5s & 160.3s & 265.9tps \\
3$\times$32B + 7$\times$14B & 96.7s &15.6s & 132.5s& 334.8tps \\
15$\times$14B & 35.4s &5.3s  & 41.9s & 754.2tps \\
\bottomrule
\end{tabular*}
\vspace{-1mm}
\end{table}

\textbf{Frameworks for Multi-Agent Systems.}
Although prior works achieve algorithm-level optimization on communication protocols \cite{zhang2025learning}, coordination strategies \cite{wang2023cooperative,qin2025strategic,han2025joyagents,liu2026learning}, and reward functions \cite{wei2025lero,su2026end,carter2026multi}, the infrastructure bottlenecks that support these MARL algorithms are overlooked. 
Regarding system-level works, some libraries (such as AutoGen \cite{wu2024autogen}, MetaGPT \cite{hong2023metagpt}, and MARFT \cite{liao2025marft}) simplify deployment through high-level abstractions without collaborative multi-agent training support. 
Dr. MAS~\cite{feng2026dr} identifies gradient-norm instability in GRPO-style multi-agent training and stabilizes optimization with agent-wise normalization.
Other MARL frameworks \cite{zhao2025stronger,wei2025multi} only provide basic training capabilities for MARL and treat the training infrastructure as a black box.  
MARTI \cite{marti2025} integrates asynchronous rollouts to enhance training efficiency, but still employs static resource allocation for different agents, leading to synchronization barriers and resource underutilization, especially when multi-agent workloads are imbalanced \cite{chen2025heterogeneous}. 
In summary, existing frameworks are incremental extensions of single-agent RL. They naively optimize the rollout or training phase and do not jointly optimize all MARL-specific challenges based on the dynamic and collaborative characteristics of MARL, thus hindering large-scale training efficiency.